\documentclass[]{fairmeta}
\usepackage{times}
\usepackage{latexsym}

\usepackage[T1]{fontenc}
\usepackage{fix-cm}
\usepackage[utf8]{inputenc}

\usepackage{microtype}

\usepackage{inconsolata}

\usepackage{graphicx}

\usepackage[utf8]{inputenc} % allow utf-8 input
\usepackage[T1]{fontenc}    % use 8-bit T1 fonts
\usepackage{hyperref}       % hyperlinks
\usepackage{url}            % simple URL typesetting
\usepackage{booktabs}       % professional-quality tables
\usepackage{amsfonts}       % blackboard math symbols
\usepackage{nicefrac}       % compact symbols for 1/2, etc.
\usepackage{microtype}      % microtypography

\usepackage{float}
\usepackage{caption}
\usepackage{siunitx} % For professional table formatting

\usepackage{tikz}
\usepackage{multirow}
\usepackage[utf8]{inputenc}
\usepackage{graphicx}
\usepackage{paracol}
\usepackage{enumitem}
\usepackage{pifont}
\newcommand{\cmark}{\ding{51}} % check mark
\newcommand{\xmark}{\ding{55}} % x mark

\usepackage{svg}
\usepackage[most]{tcolorbox}
\usepackage{multicol}
\usepackage{amsmath}

\usepackage{listings}
\usepackage{amssymb}

\usepackage{colortbl}
\usepackage{makecell}
\usepackage{lipsum}
\usepackage{array}

\definecolor{lightblue}{RGB}{220,230,245} % Soft blue
\definecolor{bluetable}{RGB}{170,200,245} % Soft blue
\definecolor{lightred}{RGB}{213, 166, 195} % Soft blue
\definecolor{navyblue}{RGB}{30,50,100}    % Navy text
\arrayrulecolor{lightblue}

\newcolumntype{L}[1]{>{\raggedright\arraybackslash}p{#1}}

\definecolor{mygreen}{RGB}{152, 199, 189}
\definecolor{myblue}{RGB}{195, 166, 213}
\definecolor{mypink}{RGB}{213, 166, 195}
\definecolor{cmarkgreen}{RGB}{0, 120, 0}
\definecolor{xmarkred}{RGB}{180, 0, 0}

\newcommand{\cmarkgreen}{{\textcolor{cmarkgreen}{\cmark}}}
\newcommand{\xmarkred}{{\textcolor{xmarkred}{\xmark}}}
\usepackage{amssymb} % For \cmark symbol

\usepackage{colortbl} % For coloring table cells
\tcbuselibrary{listingsutf8}
\usepackage{listings}

\newcommand{\BenchmarkName}{PRiSM}
\newcommand{\Agent}{PrismAgent}
\newcommand{\BenchmarkExampleNumber}{24,750}
\newcommand{\TasksNumber}{five}

\usepackage{hyperref}

\newcommand{\multimodals}{multimodal VLMs}
\newcolumntype{C}[1]{>{\centering\arraybackslash}m{#1}}
\DeclareUnicodeCharacter{2026}{\ldots}     % for …
\DeclareUnicodeCharacter{2013}{--}        % for –
\DeclareUnicodeCharacter{2014}{---}       % for —
\DeclareUnicodeCharacter{2018}{`}         % for ‘
\DeclareUnicodeCharacter{2019}{'}         % for ’
\DeclareUnicodeCharacter{201C}{``}        % for “
\DeclareUnicodeCharacter{201D}{''}        % for ”

\lstdefinestyle{mylisting}{
  language=Python,
  basicstyle=\ttfamily\scriptsize,
  breaklines=true,
  frame=single,
  columns=flexible,
  keepspaces=true,
  showstringspaces=false
}

\tcbset{
  neuripsbox/.style={
    colframe=white,               % no frame line
    colback=blue!5!white,         % very light blue background
    coltitle=blue!70!black,
    fonttitle=\bfseries,
    boxrule=0pt,                  % no border line
    sharp corners,
    leftrule=0pt,
    toprule=0pt,
    bottomrule=0pt,
    rightrule=0pt,
    enlarge left by=0mm,
    enlarge right by=0mm,
    enlarge top by=0mm,
    enlarge bottom by=0mm,
  }
}

\title{\BenchmarkName: An Agentic Multimodal Benchmark for Scientific Reasoning via Python-Grounded Evaluation}

\author{Shima Imani}
\author{Seungwhan Moon}
\author{Adel Ahmadyan}
\author{Lu Zhang}
\author{Kirmani Ahmed}
\author{Babak Damavandi}

\affiliation{Meta Reality Lab}
\abstract{Evaluating vision-language models (VLMs) in scientific domains like mathematics and physics poses unique challenges that go far beyond predicting final answers. These domains demand conceptual understanding, symbolic reasoning, and adherence to formal laws, requirements that most existing benchmarks fail to address. In particular, current datasets tend to be static, lacking intermediate reasoning steps, robustness to variations, or mechanisms for verifying scientific correctness. To address these limitations, we introduce \BenchmarkName,
a synthetic, fully dynamic, and multimodal benchmark for evaluating scientific reasoning via grounded Python code. \BenchmarkName\ includes over \BenchmarkExampleNumber\ university-level physics and math problems, and it leverages our scalable agent-based pipeline, \Agent, to generate well-structured problem instances. Each problem contains dynamic textual and visual input, a generated figure, alongside rich structured outputs: executable Python code for ground truth generation and verification, and detailed step-by-step reasoning.
The dynamic nature and Python-powered automated ground truth generation of our benchmark allow for fine-grained experimental auditing of \multimodals, revealing failure modes, uncertainty behaviors, and limitations in scientific reasoning.
To this end, we propose \TasksNumber\ targeted evaluation tasks covering generalization, symbolic program synthesis, perturbation robustness, reasoning correction, and ambiguity resolution. Through comprehensive evaluation of existing VLMs, we highlight their limitations and showcase how \BenchmarkName\ enables deeper insights into their scientific reasoning capabilities.
}

\date{\today}
\correspondence{First Author at \email{shimaimani@meta.com}}

\begin{document}

\maketitle

\section{Introduction}
\begin{figure*}[t]
    \centering
    \includegraphics[height=9.5cm]{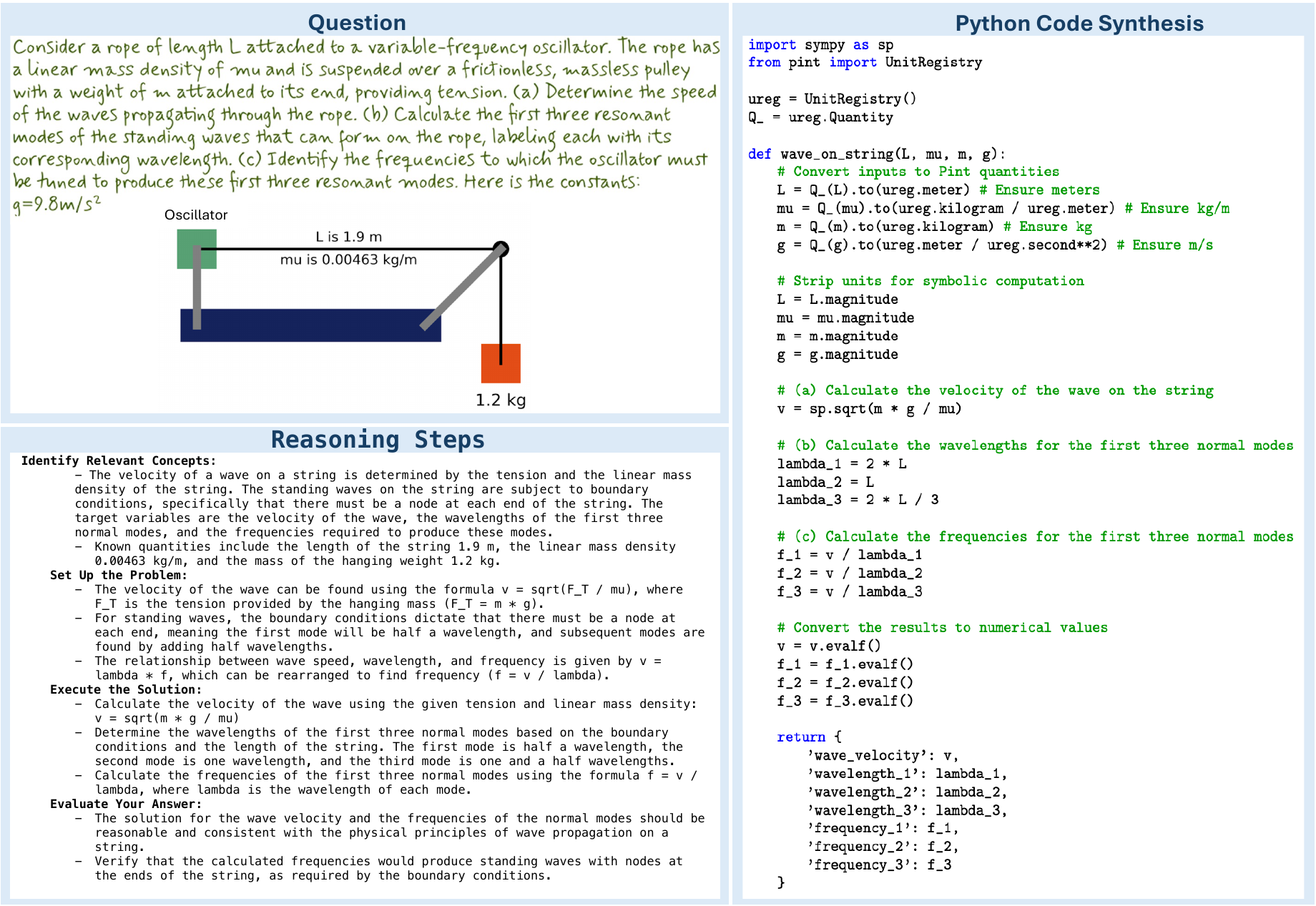}
    \caption{An instance from the \BenchmarkName\ dataset includes a parameterized question with substituted input variables, a figure generated with problem inputs, a step-by-step solution, and the corresponding Python code.}
    \label{fig:datasetexample}
\end{figure*}

\begin{figure*}[t]
    \centering
    \includegraphics[width=13cm]{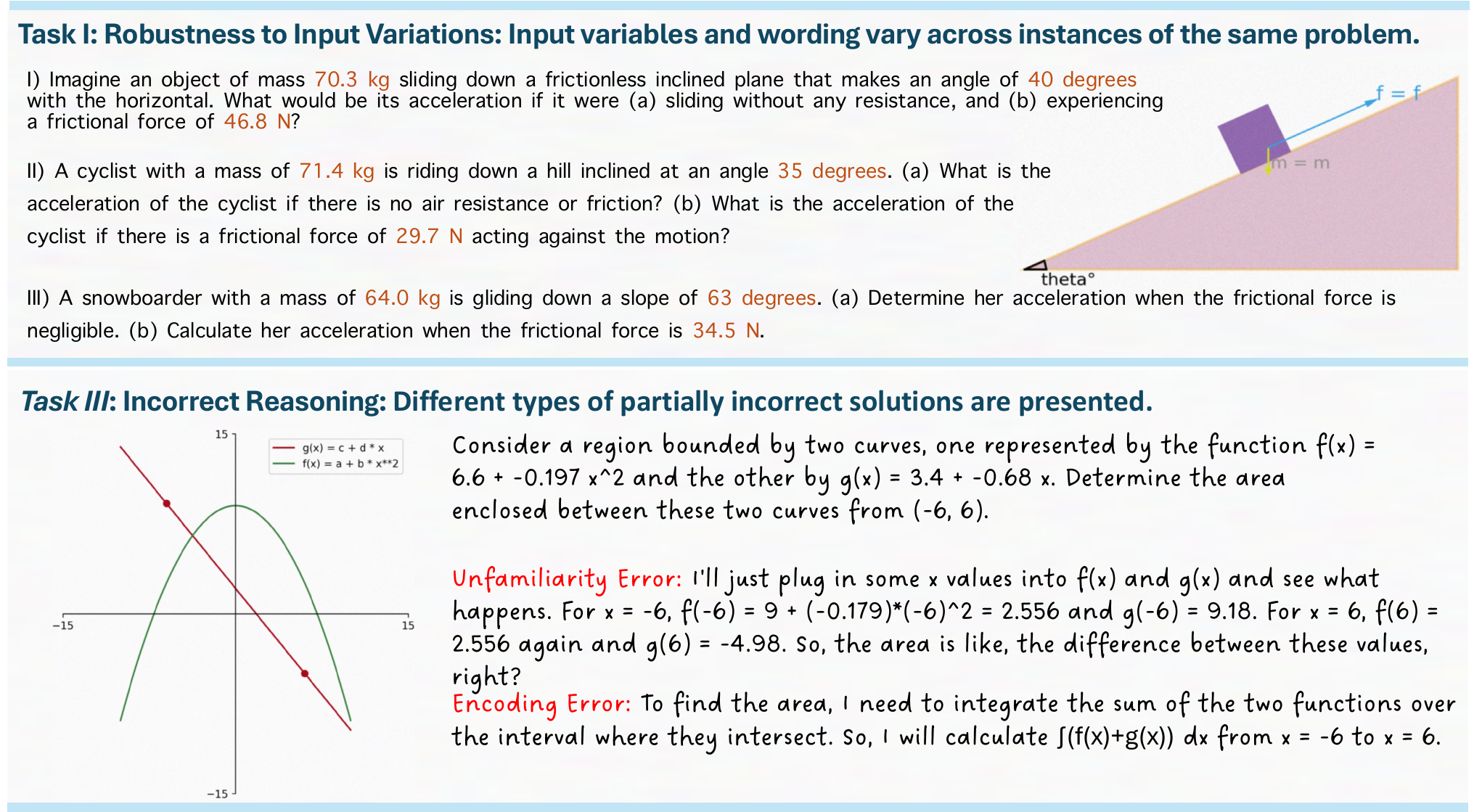}
    \caption{Examples from the \BenchmarkName~dataset illustrating Task I (Robustness to Input Variations) and Task III (Reasoning with Correction). For Task I, we vary input values and paraphrase the problem text to evaluate numerical generalization. For Task III, we introduce different types of incorrect reasoning steps and assess the model's ability to detect and correct mistakes in multi-step solutions.
}
    \label{fig:tasksexample}
\end{figure*}

Effective reasoning is fundamental for systematic problem-solving, logical deduction, and structured decision-making. In complex scientific domains like mathematics and physics, accurate reasoning requires the explicit integration of theoretical principles, rigorous mathematical processes, and computational verification to ensure dimensional and numerical validity~\cite{lake2017building, polya2014solve, chi1981categorization, newell1972human}.

Recent advancements in \multimodals~have significantly improved their reasoning capabilities. Innovations like advanced prompting techniques (e.g., chain-of-thought, tree-of-thought, and self-reflection), supervised fine-tuning on reasoning tasks, direct preference optimization (DPO), and reinforcement learning with human feedback (RLHF) have contributed to these improvements~\cite{kojima2022large,wei2022chain, yao2023tree, renze2024self,ouyang2022training, rafailov2023direct}. Alongside these advancements, a variety of datasets and benchmarks have emerged, explicitly designed to assess and enhance the reasoning skills of models in scientific contexts ~\cite{lu2022learn, wang2023scibench, hendrycks2021measuring, sun2024scieval}.

% However, despite these advancements, current datasets and benchmarks for evaluating scientific reasoning in multimodal models face several key limitations:
Nonetheless, existing benchmarks remain limited in several critical ways:

\begin{itemize}[leftmargin=*]
\item \textbf{Limited Generalization:} Most benchmarks rely on static problem formulations, lacking systematic variations or paraphrases. Furthermore, the visual modality is typically static. Consequently, these datasets fail to rigorously assess models' generalization and robustness across diverse problem settings and parameter variations.

\item \textbf{Missing Intermediate Reasoning:} The majority of datasets provide only final numerical answers, omitting detailed intermediate reasoning steps. This absence of granular guidance hinders a comprehensive evaluation of a model's reasoning path transparency, logical coherence, and interpretability.

\item \textbf{Lack of Integrated Computational Verification:} Existing benchmarks generally do not offer computational mechanisms to rigorously evaluate the quantitative correctness and dimensional consistency of solutions, crucial for scientific reasoning in mathematics and physics.
\item \textbf{Lack of Fine-Grained Error Analysis:} Existing benchmarks often focus on overall accuracy, without providing detailed breakdowns of the specific types of errors models make. This lack of detailed error categorization limits the understanding of model weaknesses and hinders targeted improvement.

\end{itemize}

To address these limitations, we present \textbf{\BenchmarkName}~(\textbf{P}ython-based \textbf{R}eason\textbf{i}ng for \textbf{S}cience and \textbf{M}ath), a fully dynamic, synthetic, and multimodal benchmark that systematically targets the identified gaps. Using our scalable agent-based pipeline, \Agent, we generate over \BenchmarkExampleNumber\ university-level mathematics and physics problem instances that cover a broad spectrum of scientific reasoning tasks. Its dynamic nature allows us to define a suite of diagnostic tasks targeting diverse dimensions of reasoning beyond final correctness. Our core contributions are as follows:

\begin{enumerate}
\item \textbf{An agentic data generation pipeline (\Agent):} A scalable pipeline that leverages autonomous agents to generate diverse scientific problems. Its modular design allows for efficient expansion to new domains and task types.

\item \textbf{A dynamic, multimodal dataset:} Each problem instance includes a parameterized textual prompt and programmatically generated figure, both tied to the same variable inputs. Paraphrased versions of the text are also included to assess linguistic robustness.

\item \textbf{Structured step-by-step solutions:} We provide symbolic derivations and detailed reasoning steps, allowing fine-grained evaluation of logical coherence and intermediate understanding.

\item \textbf{Executable Python code:} Each instance includes code that computes the ground truth solution using libraries such as SymPy (for symbolic math) and Pint (for unit consistency), supporting automated verification.

\item \textbf{Five diagnostic benchmark tasks:} Each task is designed to probe different reasoning capabilities, including generalization, symbolic program synthesis, robustness to visual perturbation, correction of flawed reasoning, and reasoning under ambiguity. Figure~\ref{fig:tasksexample} illustrates representative examples from some of these tasks.
\end{enumerate}

% \captionsetup{width=0.6\textwidth}
% \begin{figure*}[t]
%     \centering
%     \includegraphics[width=14cm]{figures/tasks_example_1.pdf}
%     \caption{Examples from the \BenchmarkName\ dataset illustrating four key tasks: (1) input variation for testing numerical generalization, (2) symbolic program synthesis to assess generalization and abstraction, (3) reasoning under ambiguity, and (4) correction of flawed reasoning.}
%     \label{fig:tasksexample}
% \end{figure*}

% \captionsetup{width=0.6\textwidth}
% \begin{figure*}[t]
%     \centering
%     \includegraphics[width=13cm]{figures/teaser_figure.pdf}
%     \caption{Examples from the \BenchmarkName~dataset illustrating Task I (Robustness to Input Variations) and Task III (Reasoning with Correction). For Task I, we vary input values and paraphrase the problem text to evaluate numerical generalization. For Task III, we introduce different types of incorrect reasoning steps and assess the model's ability to detect and correct mistakes in multi-step solutions.
% }
%     \label{fig:tasksexample}
% \end{figure*}

\section{Related Work}
\begin{table*}[ht]
\centering
% \scriptsize
\fontsize{7.5}{7.5}\selectfont
\renewcommand{\arraystretch}{1.2} % Row padding
\begin{tabular}{C{2.5cm} C{0.7cm} C{1.4cm} C{1.4cm} C{1.1cm} C{1.4cm}  C{1.4cm} C{1.4cm}}
\arrayrulecolor{lightblue}\hline
\rowcolor{lightblue}
\textbf{Dataset} & \textbf{Size} & \textbf{Level}  & \textbf{Reasoning Steps} & \textbf{Code Synthesis} & \textbf{Textual Variation (Q\&A)}  & \textbf{Numerical Variation (Q\&A)} & \textbf{Dynamic Generated Figure} \\
\hline
\makecell{ScienceQA \\ \cite{lu2022learn}} & 21,208 & Elem. \& Highschool & \xmarkred & \xmarkred & \xmarkred & \xmarkred  & \xmarkred\\
\midrule
\makecell{SciBench \\ \cite{wang2023scibench}} & 594 & Highschool & \xmarkred & \xmarkred & \xmarkred & \xmarkred  & \xmarkred\\
\midrule
\makecell{SciEval \\  \cite{sun2024scieval}} & 1,657 & Mixed & \xmarkred & \xmarkred & \xmarkred & \xmarkred  & \xmarkred\\
\midrule
\makecell{JEEBench \\ \cite{arora2023have}} & 512 & Highschool/College & \xmarkred & \xmarkred & \xmarkred & \xmarkred  & \xmarkred\\
\midrule
\makecell{MMMU \\  \cite{yue2024mmmu}} & 11,500 & College & \textbf{Partial} & \xmarkred & \xmarkred & \xmarkred  & \xmarkred\\
\midrule
\rowcolor[HTML]{E6F2F2}
% \vspace{0.1ex}
\makecell{\textbf{\BenchmarkName~(Ours)}} & \textbf{\BenchmarkExampleNumber} & \textbf{College} & \cmarkgreen & \cmarkgreen & \cmarkgreen & \cmarkgreen & \cmarkgreen  \\
\arrayrulecolor{lightblue}\hline
\end{tabular}
\vspace{0.3ex}
\caption{Comparison of scientific reasoning datasets. \BenchmarkName\ provides dynamic question variations (textual and numerical), structured reasoning, executable code, and generated figures capabilities.
}
\label{tab:benchmark_comparison}
\end{table*}

\paragraph{Scientific Reasoning and VLMs.}
Vision-language models (VLMs) have demonstrated strong capabilities in general-purpose multimodal reasoning, aided by methods such as chain-of-thought prompting~\cite{wei2022chain}, tree-of-thought~\cite{yao2023tree}, self-reflection~\cite{renze2024self}, instruction tuning~\cite{ouyang2022training}, and reinforcement learning with human feedback (RLHF)~\cite{christiano2017deep}. While these techniques significantly enhance model reasoning across many tasks, scientific domains such as mathematics and physics remain especially challenging. Solving problems in these areas often requires precise symbolic manipulation, dimensional analysis, and adherence to formal physical laws, skills that are not explicitly targeted by most existing training paradigms or benchmarks~\cite{hendrycks2021measuring, lake2017building, chi1981categorization}.

\paragraph{Benchmarks for Scientific Reasoning.}
Several datasets have been introduced to evaluate scientific reasoning. ScienceQA~\cite{lu2022learn} provides multimodal science questions aimed at grade school curricula but focuses only on final answers. SciBench~\cite{wang2023scibench} covers multiple scientific domains using text but lacks step-by-step reasoning or executable validation. MMMU~\cite{yue2024mmmu} includes college-level multimodal questions but does not offer symbolic solutions or visual perturbations. JEEBench~\cite{arora2023have} and SciEval~\cite{sun2024scieval} target standardized exams but provide minimal reasoning supervision and no programmatic verification. Overall, these benchmarks are mostly static and do not support fine-grained auditing, dynamic variation, or computational verification.

\paragraph{Executable and Tool-Augmented Reasoning.}
Recent work has shown that integrating external tools improves model performance on reasoning tasks. Program-aided reasoning frameworks such as PAL~\cite{gao2023pal}, Toolformer~\cite{schick2023toolformer}, and plugin-based environments~\cite{OpenAI2023plugins} allow models to use symbolic computation and external APIs to generate and verify answers. These frameworks highlight the value of structured program synthesis and execution for rigorous validation, an approach our benchmark adopts at scale through Python-based symbolic reasoning and verification.

\paragraph{Synthetic Benchmarks and Model Auditing.}
Synthetic data offers a controlled, scalable means to study model behavior, generalization, and failure modes~\cite{ding2021retiring, wang2024benchmark}. Dynamic synthetic benchmarks can isolate reasoning abilities and test robustness by introducing input perturbations or ambiguous cases. Despite this potential, most existing science-focused benchmarks remain static and cannot probe how VLMs respond to controlled variations. Our work builds on this motivation by offering a fully dynamic benchmark with executable ground truth, enabling fine-grained audits of multimodal reasoning in scientific domains. Additionally, we define \TasksNumber\ benchmark tasks for a comprehensive evaluation of existing VLMs.

Table~\ref{tab:benchmark_comparison} provides a detailed comparison, highlighting the distinctive features and contributions of \textsc{\BenchmarkName} in relation to existing benchmarks.

% \newcolumntype{C}[1]{>{\centering\arraybackslash}m{#1}}
% \vspace{-0.5cm}
\section{\BenchmarkName: Benchmark Overview}
\label{sec:benchmark_overview}
\subsection{Methodology}

\begin{figure*}[ht]
    \centering
    \includegraphics[height=9cm]{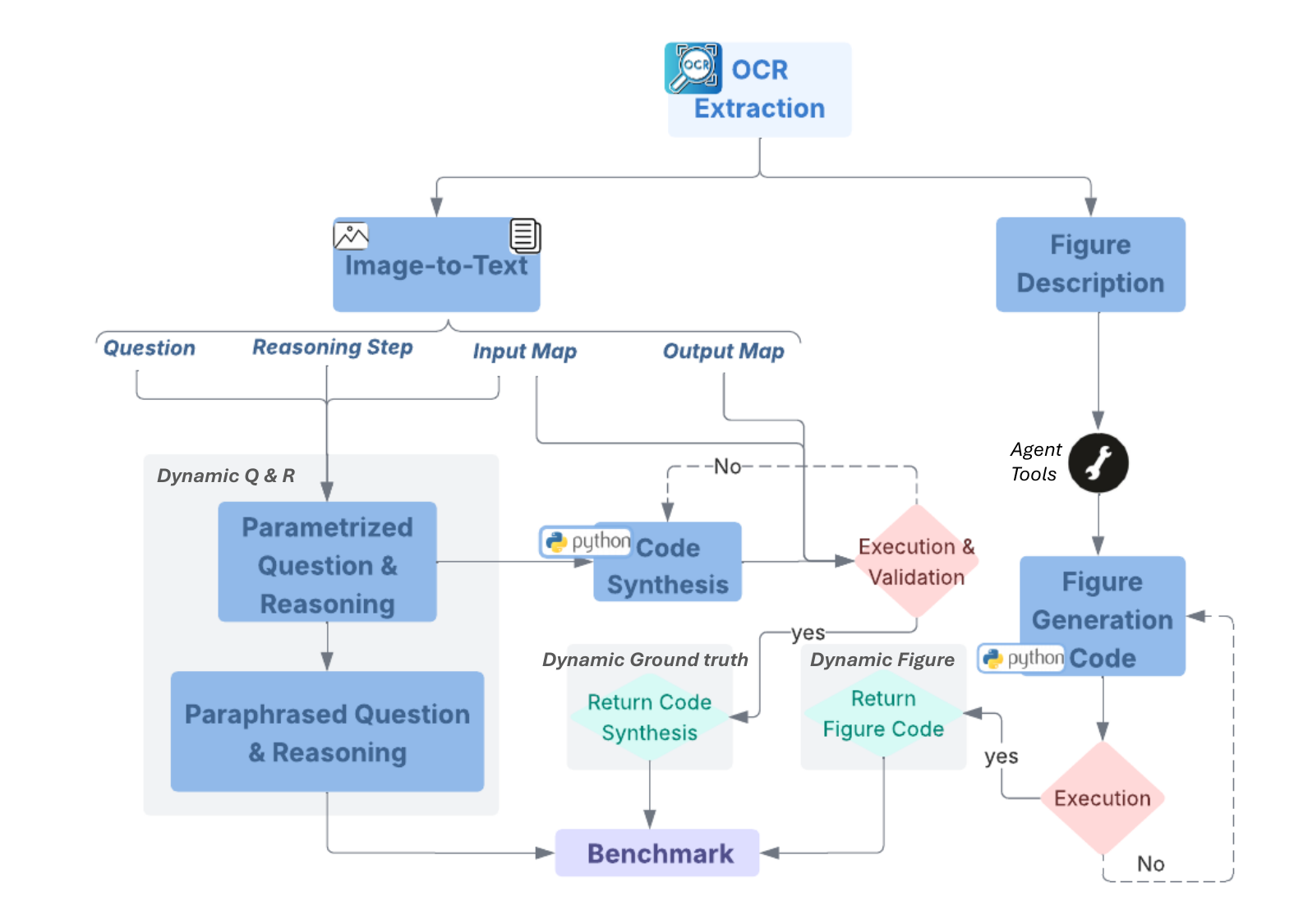}
    \caption{Overview of the Dataset Creation Pipeline Enabled by Our Framework.}
    \label{fig:chart}
\end{figure*}

In this section, we describe the construction of the agent-based pipeline \Agent~and the structure of the resulting dataset. Figure~\ref{fig:chart} provides an overview of our automated pipeline \Agent~for constructing the \BenchmarkName\ benchmark.

Importantly, all final examples were manually reviewed and filtered to include only correct and consistent instances, ensuring alignment between the question, code, and generated figure.

\subsection{OCR Extraction}

We begin by converting undergraduate-level physics and mathematics materials (primarily in PDF format) into images. All original problem sources in \BenchmarkName\ are derived from openly available Creative Commons licensed content under the CC-BY-NC license\footnote{Portions of the data were generated by LLaMA versions 3.2 and 3.3 and are subject to their respective licenses:
\url{https://github.com/meta-llama/llama-models/blob/main/models/llama3_2/LICENSE} and
\url{https://github.com/meta-llama/llama-models/blob/main/models/llama3_3/LICENSE}.}. Using Optical Character Recognition (OCR) in combination with heuristic layout analysis, we segment and crop regions containing problem statements, solutions, and relevant figures. These regions serve as inputs to our automated benchmark generation pipeline.

\subsection{\Agent: Agent for Structured Extraction}

\textbf{\textit{Image-to-Text.}}
To convert cropped regions into structured textual data, \Agent\ leverages a vision-language model (e.g., LLaMa-3.2-90B-Vision-Instruct~\cite{grattafiori2024llama}) to extract the full \textit{Question} and associated \textit{Reasoning Steps}. This preserves mathematical notation, symbols, and formatting crucial for accurate representation and downstream computation.

\Agent\ also generates an \textit{Input Map}, capturing all numerical variables and physical constants (e.g., \texttt{\{"a": [2, "dimensionless"]\}}), and an \textit{Output Map}, containing the expected final results with corresponding units (e.g., \texttt{\{"v\_a\_final": [-3, "m/s"]\}}). These structured maps enable parameterized code generation and scalable variation.

\textbf{\textit{Figure Description.}}
\Agent\ produces a step-by-step description of the extracted figure by annotating only the essential elements required to solve the problem. These annotations are abstracted to enable reconstruction of the figure using Python code, avoiding reliance on the original visual appearance and enabling dynamic figure generation.

\textbf{\textit{Parameterized Question \& Reasoning.}}
To enable abstraction and generalization, \Agent\ replaces concrete values in the question and reasoning steps with placeholders derived from the InputMap and OutputMap. This produces a standardized, symbolic representation that facilitates controlled variation and code synthesis.

\textbf{\textit{Paraphrased Question \& Reasoning.}}
To assess model robustness and linguistic generalization, \Agent\ generates paraphrased versions of the parameterized question and reasoning steps. These paraphrases retain the original logical structure and variable placeholders while varying the phrasing, context, and sentence structure. This allows for the creation of diverse yet semantically equivalent instances. Although these dynamic variations lack direct executable ground truth, this is addressed in the following step via code generation from the paraphrased reasoning.

\textbf{\textit{Code Synthesis.}}
Building on the symbolic reasoning steps and structured metadata, \Agent\ synthesizes executable Python functions that fully capture the solution logic for each problem instance. Each function accepts symbolic keys from the \textit{Input Map} as inputs and returns results mapped to the corresponding keys in the \textit{Output Map}.

To guarantee scientific validity, all code relies on established libraries: \textit{Pint} enforces unit consistency throughout computations, while \textit{SymPy} supports symbolic manipulation, algebraic simplification, and equation solving~\cite{pint-library, 10.7717/peerj-cs.103}.

This step completes the dynamic solution pipeline by enabling the generation of ground-truth answers: substituting concrete numerical values into the inputs produces the correct output for any paraphrased or parameterized instance.

\textbf{\textit{Execution and Validation.}}
To ensure scientific correctness and computational reliability, each synthesized Python function undergoes a two-stage validation process. First, dimensional validation verifies that all physical quantities adhere to correct unit consistency using the \textit{Pint} library. This step is primarily relevant for physics problems involving units, and it guarantees that operations are dimensionally coherent. For problems without units, this validation trivially passes.

Second, numerical validation executes the generated function by substituting concrete values from the \textit{Input Map}, then compares the outputs against the expected numerical results specified in the \textit{Output Map}. This step verifies that the Python code produces the correct output for the original input values of the problem. If either check fails, \Agent\ automatically flags the failure and performs iterative correction until both validations pass.

\textbf{\textit{Figure Generation.}}
To reconstruct problem diagrams, \Agent\ uses symbolic figure descriptions together with its figure generation tools, which consist of a library of predefined plotting functions (e.g., \texttt{plot\_inclined\_plane}). All figure generation functions follow a standardized format: each begins with \texttt{def generate\_figure(...)} and takes symbolic input variables as arguments, producing a rendered image saved as \texttt{figure.png}.

\textbf{\textit{Execution.}}
Each generated figure function is executed by substituting concrete values from the \textit{Input Map}, verifying that the code runs without error. The function is expected to save the rendered diagram as \texttt{figure.png}. All final examples were reviewed to ensure consistency between the question, code, and figure.

\textbf{\textit{Substitution and Instance Generation.}}
A key strength of our pipeline is its modular and dynamic structure. For any parameterized or paraphrased question, we systematically substitute values across all components, problem text, reasoning steps, solution code, and figure-generation code, to produce a fully consistent and executable problem instance. Built on a symbolic foundation, the pipeline supports the scalable generation of diverse, scientifically grounded examples. Figure~\ref{fig:datasetexample} shows an end-to-end instance produced through this process. Additional examples are provided in Appendix~\ref{appendix:examples}.

\textbf{\textit{Dataset Distribution.}}
To characterize the resulting dataset, \BenchmarkName\ comprises \BenchmarkExampleNumber\ multimodal problem instances spanning Mathematics and Physics. Table~\ref{tab:benchmark_distribution} details the distribution across domains and subtopics.
% \begin{table}[!htbp]
% \centering
% \fontsize{9}{9}\selectfont
% \renewcommand{\arraystretch}{1.1}
% \arrayrulecolor{bluetable}
% % \setlength{\heavyrulewidth}{1.2pt}
% \begin{tabular}{@{}p{4cm} p{4.5cm} p{3cm}@{}} % Three columns with adjusted widths
% \toprule
% \textbf{Mathematics (18.18\%)} & \textbf{Physics \textbf{(81.82\%)}} & \textbf{} \\
% \specialrule{1.2pt}{0pt}{0pt}  % Replace \midrule with consistent thickness
% Calculus: 7.27\% & Kinematics \& Dynamics: 36.36\% & \\
% Set Theory: 1.82\% & Work \& Energy: 10.91\% & Magnetism: 12.73\% \\
% Application of Calculus: 9.09\% & Thermodynamics: 9.09\% & Electricity: 12.73\% \\
% \bottomrule
% \end{tabular}
% % \vspace{0.3em}
% \caption{Distribution of problem instances in \BenchmarkName\ across domains and scientific subtopics.}
% \label{tab:benchmark_distribution}
% \end{table}
\begin{table}[!htbp]
\centering
\caption{Distribution of problem instances in \BenchmarkName.}
\resizebox{\columnwidth}{!}{%
\renewcommand{\arraystretch}{1.1}
\arrayrulecolor{bluetable}
\begin{tabular}{@{}p{5cm} p{4.5cm} p{4cm}@{}}
\toprule
\textbf{Mathematics (18.18\%)} & \textbf{Physics (81.82\%)} & \textbf{} \\
\specialrule{1.2pt}{0pt}{0pt}
Calculus: 7.27\% & Kinematics: 36.36\% & Magnetism: 12.73\% \\
Set Theory: 1.82\% & Work \& Energy: 10.91\% & Electricity: 12.73\% \\
Application of Calculus: 9.09\% & Thermodynamics: 9.09\% \\
\bottomrule
\end{tabular}%
}
% \caption{Distribution of problem instances in \BenchmarkName.}
\label{tab:benchmark_distribution}
\end{table}
For further details on coverage and reasoning complexity, please see Appendix~\ref{appendix:stats}.

\section{Benchmark Tasks}
\label{sec:benchmark_tasks}

Our benchmark facilitates systematic and controlled variation of problem instances to comprehensively evaluate multimodal scientific reasoning capabilities. We define \TasksNumber\ tasks , each designed to assess a specific aspect of multimodal scientific reasoning: numerical generalization, robustness to visual perturbations, reasoning
correction, programmatic generalization, and reasoning under ambiguity. This multifaceted evaluation goes beyond final-answer accuracy~\cite{yue2024mmmu, wang2023scibench, lu2022learn, ding2021retiring, wang2024benchmark}, offering deeper insight into model behavior and interpretability.  Examples of these tasks appear in Figure~\ref{fig:tasksexample} and for more examples refer to the appendix \ref{appendix:examples}.
% , with a structured task overview in Table~\ref{tab:benchmark_tasks}.
For each problem instance, we generate $N=5$ perturbed versions aligned with the task-specific variation to enable a robust and comparative evaluation.

\textbf{Task I: Robustness to Input Variations.}
This task evaluates numerical generalization by varying input values and paraphrasing the problem text, while keeping the visual presentation (e.g., handwriting, style) consistent, as illustrated in Figure~\ref{fig:tasksexample}.

\textbf{Task II: Robustness to Visual Perturbations.}
Complementary to Task I, this task probes a model's resilience to variations in the visual input, such as noise and handwriting style, while keeping the input variables constant.

\textbf{Task III: Reasoning with Correction.}
This task evaluates a model's ability to generalize by identifying and correcting errors in step-by-step reasoning. We present models with partially incorrect solutions, which are synthetically generated using LLaMA 4 Maverick~\cite{meta2024llama4} to simulate common student reasoning mistakes, as characterized in cognitive science literature~\cite{domondon2025analyzing}. The errors are categorized into four types: (i) conceptual errors (misapplication of principles), (ii) careless errors (minor arithmetic mistakes), (iii) encoding errors (misinterpretation of the problem statement), and (iv) knowledge gaps errors (responses based on guessing due to knowledge gaps). Examples of such mistakes are illustrated in Figure~\ref{fig:tasksexample}.

\textbf{Task IV: Programmatic Solution Synthesis.}
This task assesses a model's ability to generate executable Python code for solving multimodal scientific problems. Each question is presented in a generic form with input variables expressed symbolically. This formulation evaluates the model’s ability to perform symbolic reasoning and generalize solution strategies using executable Python code, assessing its proficiency in tool use via Python code~\cite{schick2023toolformer, gao2023pal, chen2021evaluating}. We execute the generated Python code on different variations of input values and compare the outputs with ground-truth answers obtained from reference Python implementations.

\textbf{Task V: Reasoning Under Ambiguity.}
This task evaluates reasoning under uncertainty, a key challenge in real-world scientific problem solving~\cite{weiss2003expressing, kahneman1982judgment}. We introduce uncertainty by systematically masking 20-30\% of input variables and replacing them with symbolic placeholders. We analyze whether models (i) defer or seek clarification, (ii) maintain symbolic representations, or (iii) make assumptions without sufficient information. These behaviors offer valuable insights into model reliability when dealing with incomplete information.

\section{Evaluation}\label{sec:evaluation_metrics}
\subsection{Evaluation Metrics}

\begin{table*}[t]
\centering
\fontsize{6.5}{7.5}\selectfont
\setlength{\tabcolsep}{2.5pt}
\caption{Performance of different models on Task I-IV across various metrics.}
\label{tab:multitask-results}
\renewcommand{\arraystretch}{1.2}
\begin{tabular}{llcccccccc}
\toprule
\textbf{Task} & \textbf{Metric} & \textbf{Qwen-72B} & \textbf{LLaMA-4 Maverick} &  \textbf{Mistral Medium-3} & \textbf{Claude-3.7 Sonnet} & \textbf{Gemini-2.5 Pro} & \textbf{GPT-4o} & \textbf{o4-mini-high} \\
 &  & \cite{qwen25} & \cite{meta2025llama4} &  \cite{mistral2025} & \cite{anthropic2024sonnet} & \cite{google2025gemini25pro} & \cite{openai2024gpt4o} & \cite{openai2025gpto3}
\\
\midrule
\multirow{4}{*}{\textbf{I}}
  & Overall Accuracy $\uparrow$    & 51.4 & 69.2 & 57.6 & 61.6 & 78.2 & 55.1 & \textbf{80.6} \\
  & TRUE score $\uparrow$    & 8.3  & 36.7 & 27.6 & 30.0 & 56.7 & 23.3 & \textbf{60.00} \\
  & Volatility $\downarrow$   & 33.3 & 18.3 & 15.5 & 28.3 & \textbf{6.7}  & 26.7 & 10.0 \\
  & TFR $\downarrow$        & 13.3 & 10.0 & 18.9 & 11.8 & 11.7 & 10.1 & \textbf{8.3} \\
\midrule
\multirow{4}{*}{\textbf{II}}
  & Overall Accuracy $\uparrow$    & 56.3 & 69.5 & 49.8 & 57.7  & \textbf{80.3} & 55.53 & 79.8 \\
  & TRUE score  $\uparrow$    & 23.3 & 48.3 & 24.1 & 28.3 & \textbf{63.3} & 30.0 & 61.7 \\
  & Volatility $\downarrow$   & 30.0 & 16.7 & 22.4 & 31.7 & \textbf{5.0}  & 26.7 & 8.3 \\
  & TFR  $\downarrow$        & 18.3 & 10.1 & 24.1 & 15.0 & \textbf{10.0} & 18.3 & 13.3 \\
\midrule
\multirow{4}{*}{\textbf{III}}
  & Overall Accuracy $\uparrow$    & 36.3 & 65.1 & 42.7 & 59.9 & \textbf{79.7} & 59.30 & \textbf{79.7} \\
  & TRUE score $\uparrow$   & 6.7 & 31.8 & 6.7 & 28.3 & \textbf{56.7} & 23.3 & 50.0\\
  & Volatility $\downarrow$  & 36.7 & 20.0 & 48.3 & 30.0 & \textbf{3.3} & 28.3 & 5.0 \\
  & TFR  $\downarrow$       & 15.0 & 11.7 & 15.0 & 10.0 & 10.2 & 15.0 & \textbf{8.3}\\
\midrule
\multirow{4}{*}{\textbf{IV}}
  & Overall Accuracy $\uparrow$    & 13.7 & 15.4 & 5.3 & 17.3 & 15.2 & 17.9 & \textbf{21.4} \\
  & TRUE score  $\uparrow$   & 8.9 & 7.1 & 8.9 & 12.5 & 10.7 & 10.7 & \textbf{19.6}\\
  & Volatility $\downarrow$  & 3.6 & 7.1 & 5.4 & \textbf{1.8} & 1.9 & 5.4 & 3.6 \\
  & TFR  $\downarrow$       & 75.1 & 78.5 & \textbf{75.0} & 78.6 & 78.6 & 76.8 & 76.8 \\
\bottomrule
\end{tabular}
\end{table*}

To evaluate the consistency and robustness of different models across problem variations, we report multiple metrics beyond standard accuracy, drawing inspiration from recent efforts such as SCORE~\cite{nalbandyan2025score}.
\\
\textbf{Overall Accuracy:} The overall percentage of correctly answered problem instances across all variations. Overall Accuracy:

{\fontsize{7pt}{-1pt}\selectfont
\[
\frac{\sum_{p \in \mathcal{P}} \sum_{v \in \mathcal{V}_p} \mathbb{I}[\hat{y}(p, v) = y(p, v)]}{\sum_{p \in \mathcal{P}} |\mathcal{V}_p|}
\]
}

where $\mathcal{P}$ is the set of all problems, $\mathcal{V}_p$ is the set of variations for problem $p$, $\hat{y}(p, v)$ is the model prediction, $y(p, v)$ is the ground truth, and $\mathbb{I}[\cdot]$ is the indicator function.

\textbf{TRUE Score (Total Reliable Understanding Evaluation):}
To assess the robustness of models in consistently solving each problem across its variations, we define the \textit{TRUE Score} as the proportion of problems for which the model achieves at least \( t\% \) accuracy across its associated variations. In our experiments, we set \( t = 90 \).

{\fontsize{7pt}{-1pt}\selectfont
\[
\text{TRUE}_{t\%} = \frac{1}{|\mathcal{P}|} \sum_{p \in \mathcal{P}} \mathbb{I}\left[
    \frac{1}{|\mathcal{V}_p|} \sum_{v \in \mathcal{V}_p} \mathbb{I}\left[\hat{y}(p, v) = y(p, v)\right] \geq t
\right] \times 100\%
\]
}

\textbf{Volatility Rate:} The fraction of problems where the model's accuracy across variations falls between 40\% and 60\%. This metric reflects the model's sensitivity to perturbations, indicating inconsistent behavior across semantically equivalent inputs. Volatility Rate:

{\fontsize{7pt}{-1pt}\selectfont
\[
\frac{1}{|\mathcal{P}|} \sum_{p \in \mathcal{P}} \mathbb{I}\left[0.4 \leq \frac{\sum_{v \in \mathcal{V}_p} \mathbb{I}[\hat{y}(p, v) = y(p, v)]}{|\mathcal{V}_p|} \leq 0.6\right] \times 100\%
\]
}

\textbf{Total Failure Rate (TFR):} The percentage of problems where the model fails to solve \emph{any} variation. A high TFR indicates fundamental reasoning gaps, where the model is unable to solve the problem, regardless of its phrasing or visualization.

{\fontsize{7pt}{-1pt}\selectfont
\[
\text{TFR} = \frac{1}{|\mathcal{P}|} \sum_{p \in \mathcal{P}} \mathbb{I}\left[\forall v \in \mathcal{V}_p, \hat{y}(p, v) \neq y(p, v)\right] \times 100\%
\]
}
% \resizebox{\columnwidth}{!}{$
% \text{TFR} = \frac{1}{|\mathcal{P}|} \sum_{p \in \mathcal{P}} \mathbb{I}\left[\forall v \in \mathcal{V}_p, \hat{y}(p, v) \neq y(p, v)\right] \times 100\%
% $}

\subsection{Experimental Results}
\label{sec:result_Experimental}

Table~\ref{tab:multitask-results} presents a quantitative evaluation of model performance across Tasks I–IV, focusing on robustness, reasoning fidelity, and code generation efficacy. Task V, addressing model behavior under ambiguous or underspecified inputs, is assessed qualitatively via an LLM-as-judge framework. Below, we highlight key insights from each task, emphasizing strengths and limitations across models.

\textbf{Tasks I-III:}
Our analysis of Tasks I (Robustness to Input Variations), II (Robustness to Visual Perturbations), and III (Reasoning with Correction) reveals consistent trends in how different models handle variations in input and reasoning.

For \textbf{Task I}, while overall accuracy suggests strong performance for models like Gemini-2.5 Pro and o4-mini-high (78–80\%), the TRUE score provides a more detailed measure of robustness. o4-mini-high demonstrates strong resilience with a TRUE score of 60\%, indicating consistent correctness across input variations for a majority of evaluated problem instances. In contrast, Qwen-72B, despite achieving an overall accuracy exceeding 50\%, exhibits a significantly lower TRUE score of approximately 8\% for Task I. This disparity underscores its vulnerability to even subtle rephrasing or variations in input variables.

Consistent with Task I, Gemini-2.5 Pro and o4-mini-high exhibit strong TRUE scores (63.33\% and 61.67\%, respectively) in \textbf{Task II}, highlighting their robustness to visual perturbations. In contrast, Mistral Medium-3 demonstrates a notable disparity between overall accuracy and TRUE score, suggesting a sensitivity to visual variations. For further insights, we conducted a detailed analysis to identify the errors and incorrect answers produced by models. These findings are discussed in detail in Appendix~\ref{appendix:error}.

\textbf{Task III} reveals a significant performance decline in Qwen-72B and Mistral Medium-3, resulting in substantially lower TRUE scores for these models. This performance disparity highlights a specific vulnerability of Qwen-72B and Mistral Medium-3 to the intricacies of the reasoning with correction task, contrasting with the more consistent performance observed in other evaluated models. In addition, we randomly selected 100 samples for detailed analysis of this task. The results can be found in Appendix~\ref{appendix:correction}.

For \textbf{Task~IV}, Programmatic Solution Synthesis, we assessed models’ ability to generate executable Python code using SI-unit inputs, as none of the models consistently handled unit conversion. High failure rates were observed across models due to (i) syntactic errors (e.g., missing imports, improperly defined parameters), and (ii) conceptual mistakes (e.g., incorrect symbolic derivations or variable substitutions). Among all models, \texttt{o4-mini-high} exhibited the most consistent and accurate performance in generating executable code. For future work, we intend to investigate these metrics after enabling the model to self-correct by providing them with the execution tool to run their code.

Finally, for \textbf{Task~V} (Ambiguity Handling), we evaluated models under incomplete or ambiguous inputs, instructing them to request clarification when necessary. Using LLaMA-4 Maverick as an LLM-based judge, we assessed whether models (i) deferred answers or asked clarifying questions, (ii) used symbolic placeholders, or (iii) made unjustified assumptions. Despite explicit instructions, models frequently ignored ambiguity, proceeding with unsupported completions. Symbolic reasoning occurred in only 4–5\% of cases, and deferral in just 3–4\%. This highlights a fundamental limitation of current VLMs, where they often prioritize providing an answer even when such an answer is not logically supportable without further clarification.
Since our judge may introduce some bias, we performed a manual review of 100 randomly selected examples from several models to verify and deepen our understanding of ambiguity handling. This qualitative analysis confirmed that models seldom ask clarifying questions and often make unsupported assumptions based on visual cues or default values, frequently using phrases such as “for simplicity, assume…”. While this enables continued problem-solving, it can mislead users into thinking these assumptions were part of the original problem. Additionally, models often produce partially simplified symbolic expressions that obscure variable dependencies. These observations reinforce the conclusion that current multimodal models tend to prioritize generating answers over managing uncertainty, underscoring the importance of developing better clarification mechanisms. Full details are provided in Appendix~\ref{appendix:ambiguity}.

% In addition, we randomly selected 100 samples for detailed analysis of this task. The results can be found in Appendix~\ref{appendix:ambiguity}.

\section{Conclusion and Future Work}
We proposed \BenchmarkName, a novel benchmark designed to rigorously evaluate scientific reasoning in vision-language models. By employing a scalable, agent-based pipeline (\Agent), we have constructed a dynamic and multimodal dataset comprising \BenchmarkExampleNumber\ university-level problems. \BenchmarkName\ is characterized by its emphasis on structured outputs, including executable Python code for ground truth generation and verification, and detailed, step-by-step reasoning. Our comprehensive evaluation, encompassing \TasksNumber\ targeted tasks, reveals significant limitations in current VLMs' ability to handle the complexities of scientific reasoning.

While current iterations of \BenchmarkName\ focus on problems within physics and mathematics, future work will broaden the scope to encompass other scientific fields and incorporate additional multimodal sources like audio and video. These extensions will augment visual reasoning challenges and enhance coverage of real-world scenarios. By expanding in these directions, \BenchmarkName\ aims to advance the rigorous assessment of VLMs and foster progress toward robust, generalizable scientific reasoning.

% \subsection{Experimental Results}
% We introduced \BenchmarkName, a benchmark for evaluating scientific reasoning in vision-language models. Built using a scalable agent-based pipeline (\Agent), it features \BenchmarkExampleNumber\ university-level, multimodal problems with structured outputs, executable code, and step-by-step reasoning. Our evaluation across \TasksNumber\ tasks highlights current VLMs' shortcomings in handling complex scientific problems. In future work, we will extend to other scientific domains and include richer modalities such as audio and video. These expansions will enhance visual reasoning challenges and support broader real-world applications, driving progress toward robust, generalizable scientific reasoning in VLMs.

\clearpage
\newpage
\bibliographystyle{assets/plainnat}
\bibliography{paper}

@article{wei2022chain,
  title={Chain-of-thought prompting elicits reasoning in large language models},
  author={Wei, Jason and Wang, Xuezhi and Schuurmans, Dale and Bosma, Maarten and Xia, Fei and Chi, Ed and Le, Quoc V and Zhou, Denny and others},
  journal={Advances in neural information processing systems},
  volume={35},
  pages={24824--24837},
  year={2022}
}

@article{rafailov2023direct,
  title={Direct preference optimization: Your language model is secretly a reward model},
  author={Rafailov, Rafael and Sharma, Archit and Mitchell, Eric and Manning, Christopher D and Ermon, Stefano and Finn, Chelsea},
  journal={Advances in Neural Information Processing Systems},
  volume={36},
  pages={53728--53741},
  year={2023}
}

@article{ouyang2022training,
  title={Training language models to follow instructions with human feedback},
  author={Ouyang, Long and Wu, Jeffrey and Jiang, Xu and Almeida, Diogo and Wainwright, Carroll and Mishkin, Pamela and Zhang, Chong and Agarwal, Sandhini and Slama, Katarina and Ray, Alex and others},
  journal={Advances in neural information processing systems},
  volume={35},
  pages={27730--27744},
  year={2022}
}

@article{christiano2017deep,
  title={Deep reinforcement learning from human preferences},
  author={Christiano, Paul F and Leike, Jan and Brown, Tom and Martic, Miljan and Legg, Shane and Amodei, Dario},
  journal={Advances in neural information processing systems},
  volume={30},
  year={2017}
}

@article{hendrycks2021measuring,
  title={Measuring mathematical problem solving with the math dataset},
  author={Hendrycks, Dan and Burns, Collin and Kadavath, Saurav and Arora, Akul and Basart, Steven and Tang, Eric and Song, Dawn and Steinhardt, Jacob},
  journal={arXiv preprint arXiv:2103.03874},
  year={2021}
}

@article{wang2023scibench,
  title={Scibench: Evaluating college-level scientific problem-solving abilities of large language models},
  author={Wang, Xiaoxuan and Hu, Ziniu and Lu, Pan and Zhu, Yanqiao and Zhang, Jieyu and Subramaniam, Satyen and Loomba, Arjun R and Zhang, Shichang and Sun, Yizhou and Wang, Wei},
  journal={arXiv preprint arXiv:2307.10635},
  year={2023}
}

@article{lu2022learn,
  title={Learn to explain: Multimodal reasoning via thought chains for science question answering},
  author={Lu, Pan and Mishra, Swaroop and Xia, Tanglin and Qiu, Liang and Chang, Kai-Wei and Zhu, Song-Chun and Tafjord, Oyvind and Clark, Peter and Kalyan, Ashwin},
  journal={Advances in Neural Information Processing Systems},
  volume={35},
  pages={2507--2521},
  year={2022}
}

@inproceedings{sun2024scieval,
  title={Scieval: A multi-level large language model evaluation benchmark for scientific research},
  author={Sun, Liangtai and Han, Yang and Zhao, Zihan and Ma, Da and Shen, Zhennan and Chen, Baocai and Chen, Lu and Yu, Kai},
  booktitle={Proceedings of the AAAI Conference on Artificial Intelligence},
  volume={38},
  pages={19053--19061},
  year={2024}
}

@article{yao2023tree,
  title={Tree of thoughts: Deliberate problem solving with large language models, 2023},
  author={Yao, Shunyu and Yu, Dian and Zhao, Jeffrey and Shafran, Izhak and Griffiths, Thomas L and Cao, Yuan and Narasimhan, Karthik},
  journal={URL https://arxiv. org/abs/2305.10601},
  volume={3},
  year={2023}
}

@article{renze2024self,
  title={Self-reflection in llm agents: Effects on problem-solving performance},
  author={Renze, Matthew and Guven, Erhan},
  journal={arXiv preprint arXiv:2405.06682},
  year={2024}
}

@article{kojima2022large,
  title={Large language models are zero-shot reasoners},
  author={Kojima, Takeshi and Gu, Shixiang Shane and Reid, Machel and Matsuo, Yutaka and Iwasawa, Yusuke},
  journal={Advances in neural information processing systems},
  volume={35},
  pages={22199--22213},
  year={2022}
}

@misc{OpenAI2023plugins,
  author       = {OpenAI},
  title        = {ChatGPT Plugins},
  year         = {2023},
  url          = {https://openai.com/index/chatgpt-plugins/},
  note         = {Accessed: 2025-03-18}
}

@article{schick2023toolformer,
  title={Toolformer: Language models can teach themselves to use tools},
  author={Schick, Timo and Dwivedi-Yu, Jane and Dess{\`\i}, Roberto and Raileanu, Roberta and Lomeli, Maria and Hambro, Eric and Zettlemoyer, Luke and Cancedda, Nicola and Scialom, Thomas},
  journal={Advances in Neural Information Processing Systems},
  volume={36},
  pages={68539--68551},
  year={2023}
}

@inproceedings{gao2023pal,
  title={Pal: Program-aided language models},
  author={Gao, Luyu and Madaan, Aman and Zhou, Shuyan and Alon, Uri and Liu, Pengfei and Yang, Yiming and Callan, Jamie and Neubig, Graham},
  booktitle={International Conference on Machine Learning},
  pages={10764--10799},
  year={2023},
  organization={PMLR}
}

@article{arora2023have,
  title={Have LLMs advanced enough? A challenging problem solving benchmark for large language models},
  author={Arora, Daman and Singh, Himanshu Gaurav and others},
  journal={arXiv preprint arXiv:2305.15074},
  year={2023}
}

@article{grattafiori2024llama,
  title={The llama 3 herd of models},
  author={Grattafiori, Aaron and Dubey, Abhimanyu and Jauhri, Abhinav and Pandey, Abhinav and Kadian, Abhishek and Al-Dahle, Ahmad and Letman, Aiesha and Mathur, Akhil and Schelten, Alan and Vaughan, Alex and others},
  journal={arXiv preprint arXiv:2407.21783},
  year={2024}
}

@article{domondon2025analyzing,
  title={Analyzing the Errors of STEM Students in Solving Basic Calculus Problems},
  author={Domondon, Christian},
  journal={Diversitas Journal},
  volume={10},
  number={1},
  year={2025}
}

@article{lake2017building,
  title={Building machines that learn and think like people},
  author={Lake, Brenden M and Ullman, Tomer D and Tenenbaum, Joshua B and Gershman, Samuel J},
  journal={Behavioral and brain sciences},
  volume={40},
  pages={e253},
  year={2017},
  publisher={Cambridge University Press}
}

@incollection{polya2014solve,
  title={How to solve it: A new aspect of mathematical method},
  author={Polya, George},
  booktitle={How to solve it},
  year={2014},
  publisher={Princeton university press}
}

@article{chi1981categorization,
  title={Categorization and representation of physics problems by experts and novices},
  author={Chi, Michelene TH and Feltovich, Paul J and Glaser, Robert},
  journal={Cognitive science},
  volume={5},
  number={2},
  pages={121--152},
  year={1981},
  publisher={Elsevier}
}

@article{newell1972human,
  author  = {Newell, Allen and Simon, Herbert},
  title   = {Human Problem Solving},
  journal = {Prentice-Hall},
  year    = {1972},
  volume  = {1}, 
}

@article{10.7717/peerj-cs.103,
 title = {SymPy: symbolic computing in Python},
 author = {Meurer, Aaron and Smith, Christopher P. and Paprocki, Mateusz and \v{C}ert\'{i}k, Ond\v{r}ej and Kirpichev, Sergey B. and Rocklin, Matthew and Kumar, AMiT and Ivanov, Sergiu and Moore, Jason K. and Singh, Sartaj and Rathnayake, Thilina and Vig, Sean and Granger, Brian E. and Muller, Richard P. and Bonazzi, Francesco and Gupta, Harsh and Vats, Shivam and Johansson, Fredrik and Pedregosa, Fabian and Curry, Matthew J. and Terrel, Andy R. and Rou\v{c}ka, \v{S}t\v{e}p\'{a}n and Saboo, Ashutosh and Fernando, Isuru and Kulal, Sumith and Cimrman, Robert and Scopatz, Anthony},
 year = 2017,
 month = jan,
 volume = 3,
 pages = {e103},
 journal = {PeerJ Computer Science},
 issn = {2376-5992},
 url = {https://doi.org/10.7717/peerj-cs.103},
 doi = {10.7717/peerj-cs.103}
}

@misc{pint-library,
  title        = {Pint: Python Units Library},
  author       = {{Pint Developers}},   
  year         = {2025},
  url          = {https://pint.readthedocs.io/},
}

@inproceedings{yue2024mmmu,
  title={Mmmu: A massive multi-discipline multimodal understanding and reasoning benchmark for expert agi},
  author={Yue, Xiang and Ni, Yuansheng and Zhang, Kai and Zheng, Tianyu and Liu, Ruoqi and Zhang, Ge and Stevens, Samuel and Jiang, Dongfu and Ren, Weiming and Sun, Yuxuan and others},
  booktitle={Proceedings of the IEEE/CVF Conference on Computer Vision and Pattern Recognition},
  pages={9556--9567},
  year={2024}
}

@article{ding2021retiring,
  title={Retiring adult: New datasets for fair machine learning},
  author={Ding, Frances and Hardt, Moritz and Miller, John and Schmidt, Ludwig},
  journal={Advances in neural information processing systems},
  volume={34},
  pages={6478--6490},
  year={2021}
}

@article{wang2024benchmark,
  title={Benchmark suites instead of leaderboards for evaluating AI fairness},
  author={Wang, Angelina and Hertzmann, Aaron and Russakovsky, Olga},
  journal={Patterns},
  volume={5},
  number={11},
  year={2024},
  publisher={Elsevier}
}

@misc{meta2024llama4,
  author       = {Meta AI},
  title        = {LLaMA 4: Advancing Open Multimodal Intelligence},
  year         = {2024},
  howpublished = {\url{https://ai.meta.com/blog/llama-4-multimodal-intelligence/}},
  note         = {Accessed: 2025-05-10}
}

@article{weiss2003expressing,
    author = {Weiss, Charles},
    title = {Expressing scientific uncertainty},
    journal = {Law, Probability and Risk},
    volume = {2},
    number = {1},
    pages = {25-46},
    year = {2003},
    month = {03},
    issn = {1470-8396},
    doi = {10.1093/lpr/2.1.25},
    url = {https://doi.org/10.1093/lpr/2.1.25},
    eprint = {https://academic.oup.com/lpr/article-pdf/2/1/25/2777585/mgg004.pdf},
}

@book{kahneman1982judgment,
  title={Judgment under uncertainty: Heuristics and biases},
  author={Kahneman, Daniel and Slovic, Paul and Tversky, Amos},
  year={1982},
  publisher={Cambridge university press}
}

@article{nalbandyan2025score,
  title={SCORE: Systematic COnsistency and Robustness Evaluation for Large Language Models},
  author={Nalbandyan, Grigor and Shahbazyan, Rima and Bakhturina, Evelina},
  journal={arXiv preprint arXiv:2503.00137},
  year={2025}
}

@article{chen2021evaluating,
  title={Evaluating large language models trained on code},
  author={Chen, Mark and Tworek, Jerry and Jun, Heewoo and Yuan, Qiming and Pinto, Henrique Ponde De Oliveira and Kaplan, Jared and Edwards, Harri and Burda, Yuri and Joseph, Nicholas and Brockman, Greg and others},
  journal={arXiv preprint arXiv:2107.03374},
  year={2021}
}

@article{qwen25,
    title   = {Qwen2.5 Technical Report}, 
    author  = {An Yang and Baosong Yang and Beichen Zhang and Binyuan Hui and Bo Zheng and Bowen Yu and Chengyuan Li and Dayiheng Liu and Fei Huang and Haoran Wei and Huan Lin and Jian Yang and Jianhong Tu and Jianwei Zhang and Jianxin Yang and Jiaxi Yang and Jingren Zhou and Junyang Lin and Kai Dang and Keming Lu and Keqin Bao and Kexin Yang and Le Yu and Mei Li and Mingfeng Xue and Pei Zhang and Qin Zhu and Rui Men and Runji Lin and Tianhao Li and Tingyu Xia and Xingzhang Ren and Xuancheng Ren and Yang Fan and Yang Su and Yichang Zhang and Yu Wan and Yuqiong Liu and Zeyu Cui and Zhenru Zhang and Zihan Qiu},
    journal = {arXiv preprint arXiv:2412.15115},
    year    = {2024}
}

@misc{meta2025llama4,
  author       = {Meta AI},
  title        = {{Introducing LLaMA 4: Advancing Multimodal Intelligence}},
  howpublished = {\url{https://ai.meta.com/blog/llama-4-multimodal-intelligence/}},
  note         = {Accessed: 2025-05-12},
  year         = {2025},
  month        = apr
}

@misc{mistral2025,
  author       = {Mistral AI},
  title        = {{Mistral-medium-3: Medium is the new large}},
  howpublished = {\url{https://mistral.ai/news/mistral-medium-3/}},
  note         = {Accessed: 2025-05-12},
  year         = {2025},
  month        = apr
}

@misc{anthropic2024sonnet,
  author       = {Anthropic},
  title        = {{Introducing the Claude 3 Model Family}},
  howpublished = {\url{https://www.anthropic.com/news/claude-3-family}},
  note         = {Launch post for Claude 3 Sonnet; accessed 2025‑05‑12},
  year         = {2024},
  month        = mar
}

@misc{google2025gemini25pro,
  author       = {Google DeepMind},
  title        = {{Gemini 2.5: Our Most Intelligent AI Model}},
  howpublished = {\url{https://blog.google/technology/google-deepmind/gemini-model-thinking-updates-march-2025/}},
  note         = {Gemini 2.5 Pro announcement; accessed 2025‑05‑12},
  year         = {2025},
  month        = mar
}

@misc{openai2024gpt4o,
  author       = {OpenAI},
  title        = {Hello GPT-4o},
  howpublished = {\url{https://openai.com/index/hello-gpt-4o/}},
  note         = {Model launch page; accessed 2025-05-12},
  year         = {2024},
  month        = may
}

@misc{openai2025gpto3,
  author       = {OpenAI},
  title        = {Model Release Notes -- OpenAI o3},
  howpublished = {\url{https://help.openai.com/en/articles/9624314-model-release-notes}},
  note         = {Release notes section "OpenAI o3 and o4-mini"; accessed 2025-05-12},
  year         = {2025},
  month        = apr
}

\clearpage
\newpage
\beginappendix

\section{Analysis: Robustness to Text and Visual Perturbations (Task I \& II}\label{appendix:error}

To gain deeper insight into model failure modes within \BenchmarkName, we randomly sampled 500 examples and conducted a thorough qualitative analysis of the model outputs. Although our initial experiments focused on the output of Maverick model, we observed similar behaviors across other models such as Gemini-2.5 Pro and Qwen-72B, indicating these issues are broadly representative.

Below, we summarize the key failure categories:
\begin{figure*}[t]
    \centering
\includegraphics[width=0.8\linewidth]{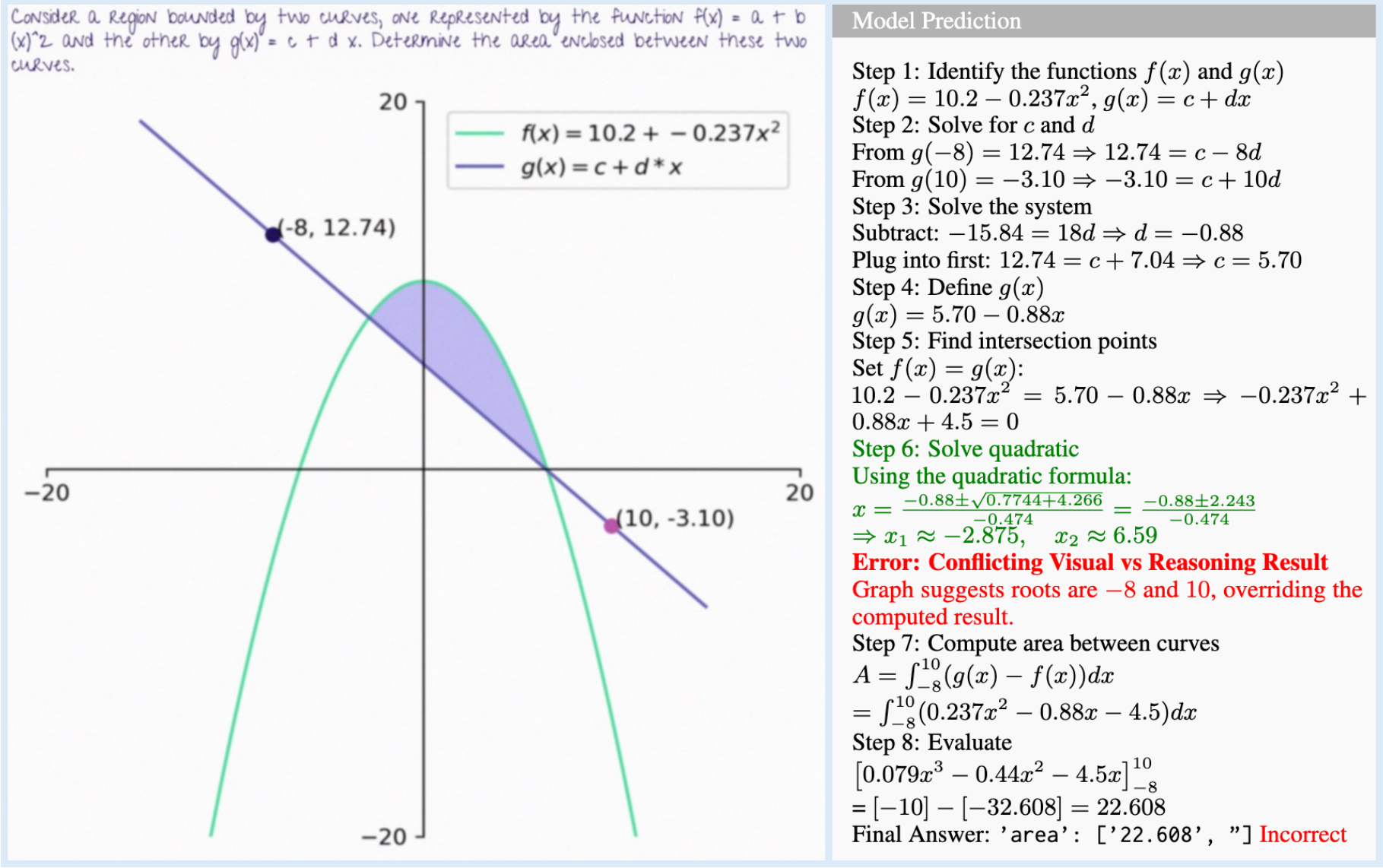}
    \caption{Illustration of modality conflict where the model’s correct algebraic solution is overridden by a misleading visual cue.}
    \label{fig:ambiguity_conflict}
\end{figure*}
\paragraph{Modality Conflict}

Despite correctly solving problems through algebraic or symbolic reasoning, the model sometimes overrides these solutions due to ambiguity or noise present in the visual modality. This reflects a lack of a robust verification mechanism to reconcile the perceptual (image-based) input with the symbolic reasoning process. As a result, the model often places undue trust in the visual information, even when it contradicts more precise calculations.

Figure \ref{fig:ambiguity_conflict} illustrates an example of this issue, with the model’s prediction displayed in the gray box. Although the model correctly computes the intersection points of two functions algebraically, it ultimately produces incorrect results by relying on a misleading visual interpretation of the graph.

\paragraph{Ambiguity-Induced Errors}

These errors arise from ambiguous or unclear visual features, such as misreading handwritten digits or overlooking critical decimal points. This issue stems from insufficient robustness in interpreting noisy or poorly rendered visual inputs, leading to incorrect numeric or symbolic interpretations that are not reconciled with the broader problem context.

\begin{figure}[h]
    \centering
    \includegraphics[width=0.5\linewidth]{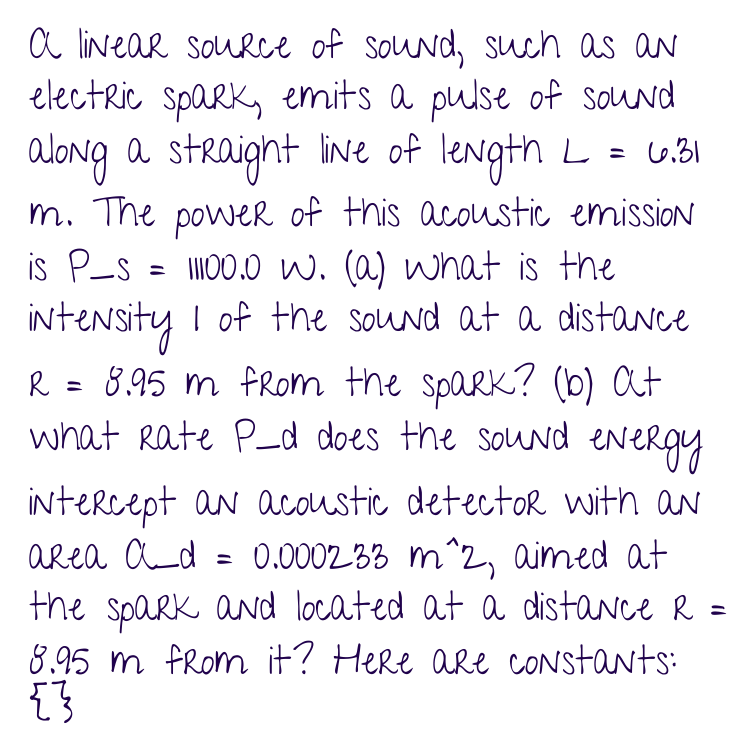}
    \caption{Example of ambiguity in handwritten digits leading to misinterpretation.}
    \label{fig:ambigous_1}
\end{figure}

As shown in Figure~\ref{fig:ambigous_1}, the handwritten value ``\(L = 6.31\) m'' was initially misread as ``\(0.31\) m'' due to a curled digit and a faint decimal point, leading the model to an incorrect interpretation. However, when presented with a zoomed-in version of the image, the model recognized and corrected the error.

\paragraph{Visual Misreading Errors}

These errors occur when the model misreads numerical values, especially in dense annotations or when dealing with large numbers.

\begin{figure}[h]
    \centering
    \includegraphics[width=0.5\linewidth]{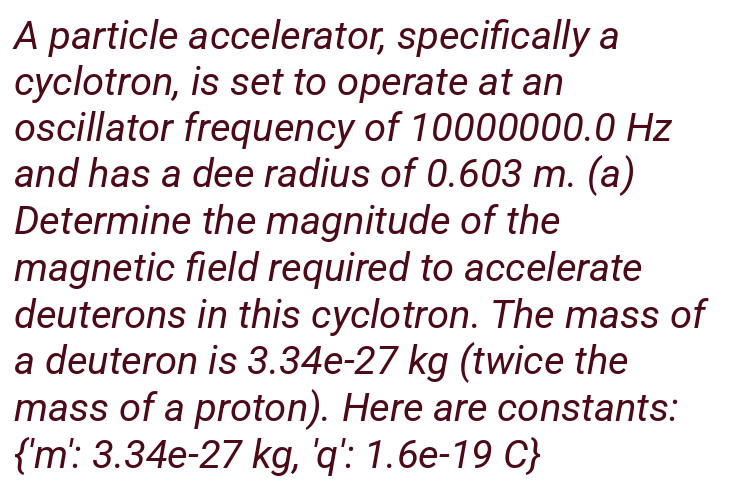}
    \caption{Example of ambiguous digit rendering leading to incorrect frequency interpretation.}
    \label{fig:visual_understanding}
\end{figure}

For example, in the question shown in Figure~\ref{fig:visual_understanding}, the model interpreted the oscillator frequency as \(f = 1{,}000{,}000.0\) Hz, whereas the correct value was \(f = 10{,}000{,}000.0\) Hz. The additional zero was visually subtle, and the model failed to verify the magnitude against contextual cues.

\vspace{2 mm}
\paragraph{Multi-Step Reasoning Gaps}

Even when the model performs step-by-step reasoning, it often fails in problems requiring decomposition of a single step into multiple intermediate hops. Attempting to collapse these multi-hop logical steps into one frequently causes the model to skip key transformations, particularly during unit conversions, chained exponential operations, or algebraic simplifications, resulting in incorrect final answers.

For example, in the following calculation, the model initially provides clear, step-by-step reasoning. However, instead of maintaining this detailed approach, it abruptly collapses the multi-hop process, such as computing exponentials and multiplications, into a single step and outputs a numerical answer without explanation.

\begin{tcolorbox}[title=Model Prediction, colback=gray!5, colframe=black!30, boxrule=0.5pt, arc=2mm]
\small

Step 3: Perform the calculation for \(K\) \\[4pt]
\[
K = \frac{1}{4\pi \times 8.85 \times 10^{-12}} \cdot \frac{(1.6 \times 10^{-19})^2}{2 \times 10^{-15}}
\]
\[
= \frac{1}{4\pi \times 8.85 \times 10^{-12}} \cdot \frac{2.56 \times 10^{-38}}{2 \times 10^{-15}}
\]
\[
= \frac{1}{1.113 \times 10^{-10}} \cdot 1.28 \times 10^{-23}
\]

\vspace{4pt}
\textcolor{red}{
\textbf{Incorrect:} \\
\(K = 8.99 \times 10^{-14} \cdot 1.28 \times 10^{-23} = 1.15 \times 10^{-36} \, \text{J}\)
}

\vspace{4pt}
\textcolor{green!50!black}{
\textbf{Expected:} \\
\(K = 8.99 \times 10^{9} \cdot 1.28 \times 10^{-23} = 1.15 \times 10^{-13} \, \text{J}\)
}

\vspace{6pt}
\textcolor{red}{
\textbf{Error Explanation:} The model collapses multiple steps, approximating \(1 / (1.113 \times 10^{-10})\), converting it to the Coulomb constant \(8.99 \times 10^{9}\), and then performing the multiplication, all in a single step. In doing so, it incorrectly substitutes \(8.99 \times 10^{-14}\) instead of \(8.99 \times 10^{9}\), leading to a drastic exponent error and an incorrect final result.
}

\end{tcolorbox}

\paragraph{Numerical Precision}

Models often perform numerically correct computations but fail to adjust precision appropriately for the problem context. Unlike humans, who infer the number of significant digits or reason about precision based on the physical setup, models frequently over approximate or under approximate without justification. This leads to issues such as numerical instability when subtracting large, nearly equal values (e.g., computing \(\Delta r = r_1 - r_2\) where \(r_1 \approx r_2\)), resulting in inaccurate final results despite correct symbolic expressions. Models also make errors in exponentiation calculations, for instance, computing \(1.79^5 \) and incorrectly reporting \(24.76\) where the correct answer is \(18.37\).

\paragraph{Conceptual or Formula Misapplication}

This category includes errors such as using incorrect formulas. For example, in physics problem applying \( M = \mu_0 N_1 N_2 R^2 / l \) instead of the correct \( M = \mu_0 N_1 N_2 A / l \) or misapplying the Hall voltage formula by using the wrong variable for thickness or length. Models also confuse physical setups, such as assuming standing wave modes for a string fixed at both ends when it is actually fixed at only one end. Additionally, there are mistakes in interpreting circuits, like mixing up series and parallel configurations despite visual cues.

\section{Analysis: Reasoning with Correction (Task III)}
\label{appendix:correction}

To better understand model failure modes on the \textit{Reasoning with Correction} task, we conducted a qualitative analysis of 100 randomly sampled examples output from Qwen-72B and Gemini-2.5 Pro. We categorized the models' responses into three main types:

\begin{enumerate}
    \item \textbf{Silent Correction:} The model ignores the student's error and provides a corrected solution without referencing the mistake.

    \textit{Examples:}
    \begin{itemize}
        \item ``Let's continue solving the problem step by step...''
        \item ``Ignore the student's work and start from scratch.''
    \end{itemize}

    \item \textbf{Uncritical Acceptance:} The model accepts the student's reasoning without verifying correctness.

    \textit{Examples:}
    \begin{itemize}
        \item ``The student's approach is correct. We need to complete the calculations.''
        \item ``The student calculated electric force correctly but hasn't completed the work and energy part.''
    \end{itemize}

    \item \textbf{Explicit Error Identification and Correction:} The model identifies the student's error, explains it, and then provides the corrected solution.

    \textit{Examples:}
    \begin{itemize}
        \item ``The student made a mistake in applying Ohm's law...''
        \item ``There is an algebraic error in the student's solution. Here is the corrected version...''
    \end{itemize}
\end{enumerate}

Table \ref{tab:correction_behaviors} summarizes the results of our quantitative analysis. Gemini-2.5 Pro demonstrates a notably higher propensity to explicitly detect and correct errors compared to Qwen-72B. In addition, other behaviors such as silent correction and uncritical acceptance indicate missed opportunities for offering valuable, instructive feedback. Therefore, this task evaluates not only the accuracy of model responses but also their diagnostic reasoning and ability to provide meaningful guidance.

\begin{table}[h!]
\centering
\small
\resizebox{10cm}{!}{%
\renewcommand{\arraystretch}{1.1}
\begin{tabular}{lcc}
\toprule
\textbf{Behavior Type} & \textbf{Qwen-72B} & \textbf{Gemini-2.5 Pro} \\
\midrule
Explicit Correction & 62\% & 84\% \\
Silent Correction & 24\% & 6\% \\
Uncritical Acceptance & 14\% & 10\% \\
\bottomrule
\end{tabular}
}
\caption{Distribution of observed behaviors in the Reasoning with Correction task. Note that high explicit correction rates do not guarantee correctness of the final answer.}
\label{tab:correction_behaviors}
\end{table}

\section{Analysis: Ambiguity Handling (Task  V)}\label{appendix:ambiguity}

To better understand the task and common failure modes of ambiguity task, we manually reviewed 100 random samples from OpenAI’s \texttt{o4-mini-high}, Gemini-2.5 Pro (closed-source models), and Qwen-72B (open-source model). Our qualitative analysis highlights why models rarely ask clarifying questions, even when explicitly prompted:

\begin{enumerate}
    \item \textbf{Unjustified Assumptions from Visual Context.} Models often infer missing parameters directly from the image instead of requesting clarification. For example, in a question missing horizontal coordinates \(x_1\) and \(x_2\), Qwen-72B incorrectly assumed \(x_1 = x_2\), relying only on vertical separation cues in the figure, resulting in an invalid assumption.

    \item \textbf{Assumptions Without Explicit Clarification.} When unable to visually infer values, models proceed with default assumptions, using phrases such as “for simplicity, assume...”. While this enables continued problem-solving, it can mislead users into thinking these assumptions were part of the original problem.

    \item \textbf{Partial or Incomplete Symbolic Simplifications.} Models like Gemini-2.5 Pro and \texttt{o4-mini-high} sometimes stop at symbolic answers involving unsimplified variables, which, although not incorrect, obscure the final dependency structure. For instance, if the target is \(y\) and the model outputs \(y = \frac{A}{c}\), but \(A = a \times c\), the fully simplified answer should be \(y = a\). Stopping early yields a less informative and potentially misleading result.
\end{enumerate}

Models tend to infer or assume missing information rather than ask clarifying questions, reflecting a core limitation where multimodal models prioritize solution completion over managing uncertainty. Our findings underscore the need for explicit deferral or clarification capabilities in VLMs.

\section{Predefined Plotting Functions}

Our benchmark utilizes a set of modular Python plotting utilities (mainly based on \texttt{matplotlib}) to standardize figure generation in math and physics problems. These functions, collected in \texttt{figure\_utils.py}, abstract low-level drawing, enabling the agent to produce consistent and accurate figures efficiently.

Representative functions include:
\begin{itemize}
    \item \texttt{add\_inclined\_surface\_with\_angle(ax, ...)}
    \item \texttt{add\_block\_on\_incline(ax, ...)}
    \item \texttt{draw\_horizontal\_resistor(ax, ...)}
\end{itemize}

Development was iterative, driven by dataset analysis and aided by LLMs to identify common diagram components. All generated figure code was manually reviewed for correctness.

\section{Appendix: Benchmark Statistics}\label{appendix:stats}

\paragraph{Domain Coverage} The physics domain covers 450 concepts, with core topics such as energy (2.67\%), force (2.00\%), velocity (2.00\%), and kinetic energy (1.78\%). Additional frequent topics include acceleration, mass, electric field, momentum, work, magnetic field, potential energy, torque, Coulomb force, power, and wave motion. The long tail features concepts like gravitational effects, moment of inertia, Faraday’s law, projectile motion, and lens equation (each approximately 0.22\%).

In the mathematics domain, which includes 110 concepts, top concepts are area (5.45\%), function (5.45\%), and interval (3.64\%). Other covered topics include slope, algebraic expressions, equations, symmetry, triangles, circles, inequalities, coordinate geometry, ratios, integration, differentiation, Venn diagrams, and set theory.

\paragraph{Reasoning Complexity:} Problems in \BenchmarkName\ require multi-hop, structured reasoning as reflected in the problem step counts: mean steps per problem is 6.91, with a minimum of 4 and maximum of 11 steps, and a standard deviation of 1.75. This indicates that our benchmark consistently demands deep reasoning rather than shortcut heuristics.

\paragraph{Necessity and Role of Images}
We performed a detailed analysis of how variables are distributed across the different modalities, including text and figures, within our dataset to understand the unique contributions of each modality and to demonstrate the necessity of multimodal reasoning.

\begin{itemize}
    \item \textbf{Problem Inputs:} Number of input variables required to solve the problem (from all modalities).
    \item \textbf{Problem Outputs:} Number of output variables or answers expected for the problem.
    \item \textbf{Figure Inputs:} Number of variables present in the figure (including those also in text).
    \item \textbf{Text Inputs:} Number of variables present in the text (including those also in figures).
    \item \textbf{Figure-only Inputs:} Number of variables appearing exclusively in the figure, not in text.
    \item \textbf{Text-only Inputs:} Number of variables appearing exclusively in the text, not in figures.
\end{itemize}

 \begin{table}[h]
    \centering
    \small % or \footnotesize for even smaller text
    \setlength{\tabcolsep}{4pt} % reduce horizontal padding between columns
    \begin{tabular}{@{}lcccc@{}} % remove extra space on left and right
        \toprule
        \textbf{Variable Type} & \textbf{Min} & \textbf{Max} & \textbf{Mean} & \textbf{Std Dev} \\
        \midrule
        Problem Inputs         & 1.0 & 12.0 & 4.43 & 1.65 \\
        Problem Outputs        & 1.0 & 8.0  & 2.20 & 1.81 \\
        Figure Inputs          & 1.0 & 8.0  & 3.50 & 1.40 \\
        Text Inputs            & 0.0 & 11.0 & 3.39 & 1.74 \\
        Figure-only Inputs     & 0.0 & 8.0  & 0.62 & 1.50 \\
        Text-only Inputs       & 0.0 & 9.0  & 0.52 & 1.33 \\
        \bottomrule
    \end{tabular}
    \caption{Descriptive statistics of variable counts across modalities in \BenchmarkName.}
    \label{tab:modality-vars}
\end{table}

A substantial portion of problems require information extracted jointly from both figures and text, highlighting the complementary roles of these modalities. Importantly, a significant number of variables appear \emph{exclusively} in the figures, underscoring the indispensable role of visual reasoning. Similarly, certain variables occur only in the textual modality, validating the necessity for effective multimodal fusion to support comprehensive model reasoning. Additionally, due to the dynamic nature of our benchmark, overlapping variables may be included in any modality as needed.

\section{Appendix: Examples from \BenchmarkName~dataset}
\label{appendix:examples}
In this section, we provide more examples from the \BenchmarkName~dataset.

\begin{figure*}[t]
    \centering
\includegraphics[width=0.95\textwidth,keepaspectratio]{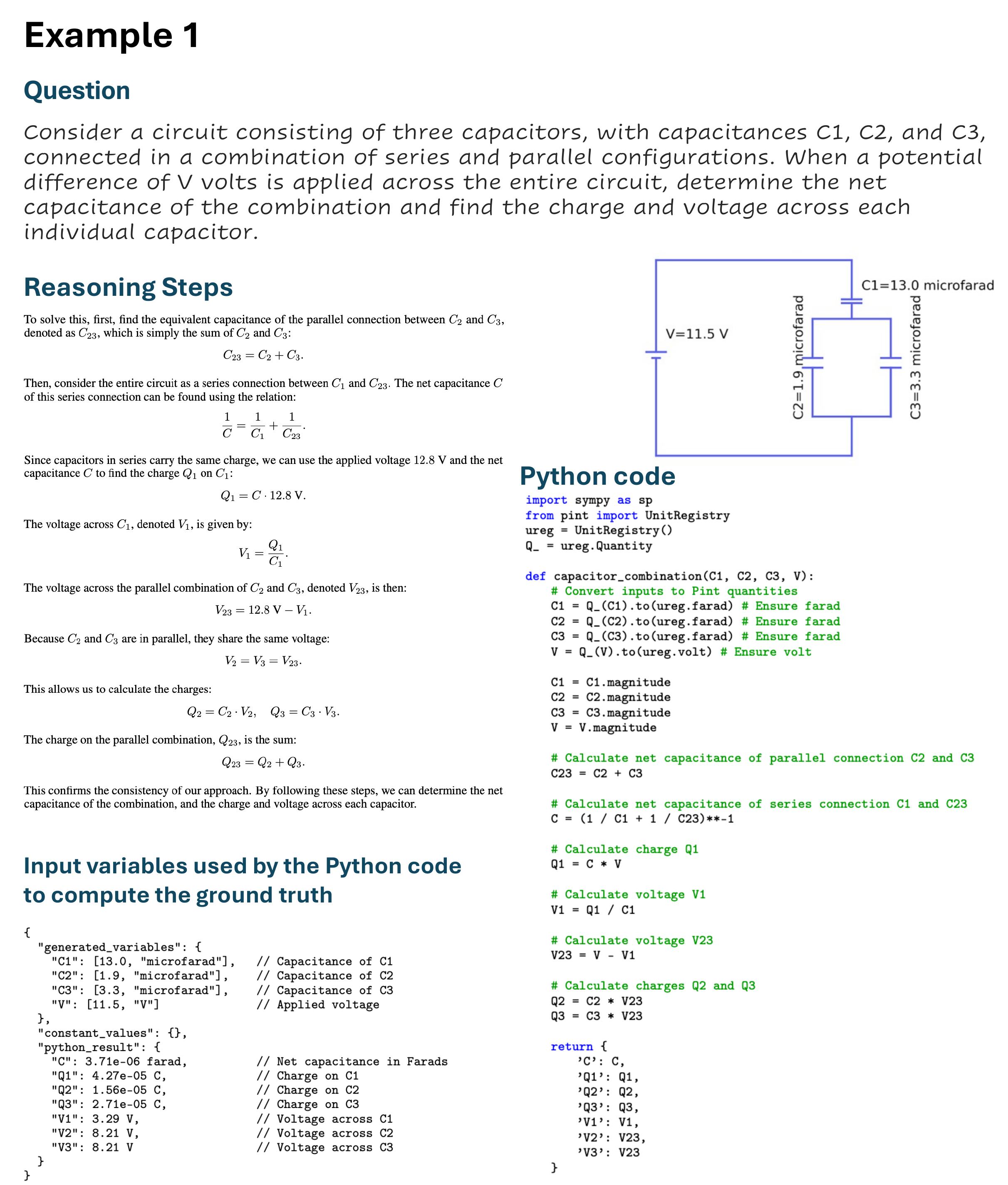}
    \caption{Example 1 from \BenchmarkName~showing input variation and structured reasoning.}
    \label{fig:example1}
\end{figure*}

\begin{figure*}[t]
    \centering
    \includegraphics[width=0.95\textwidth,keepaspectratio]{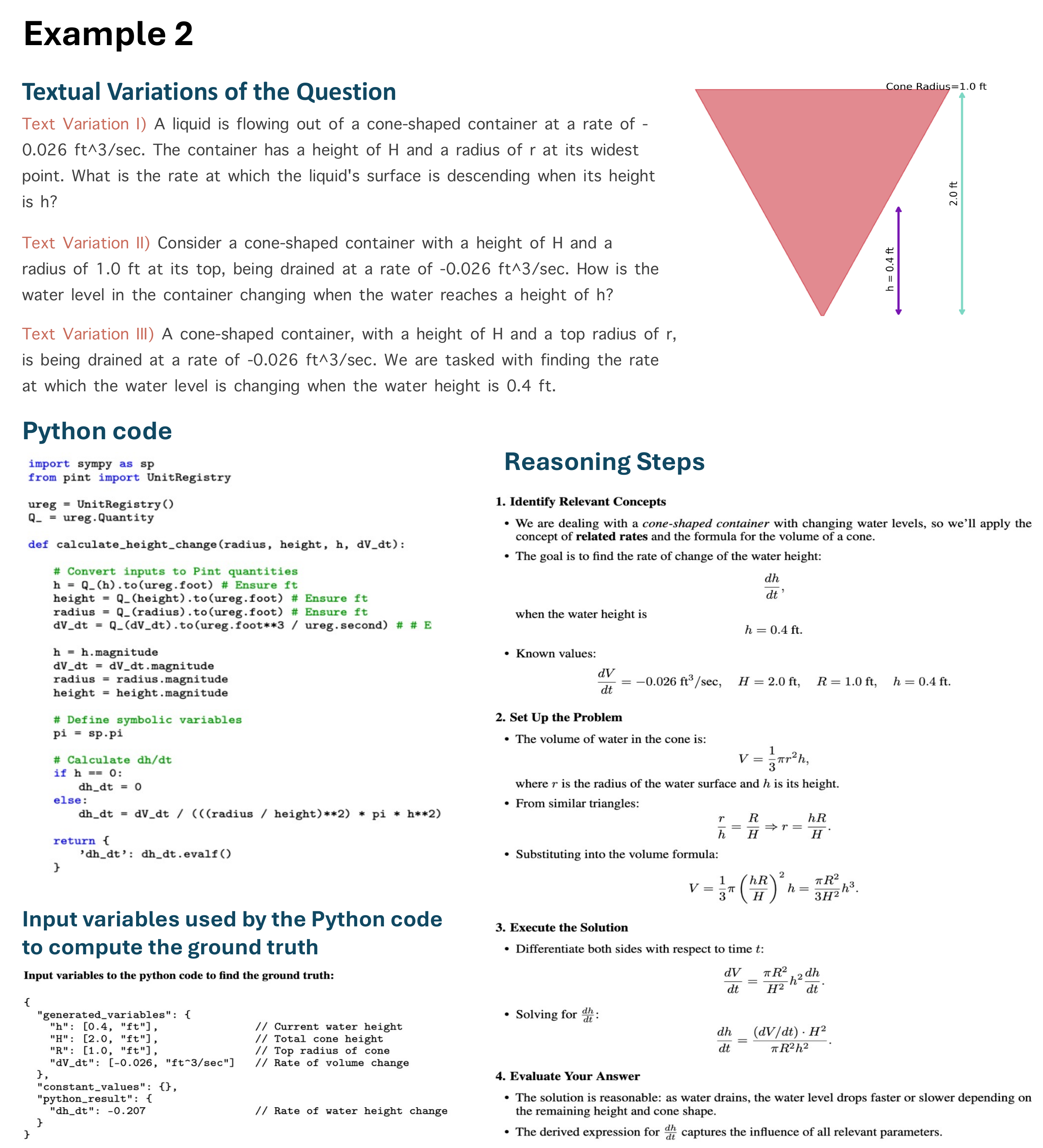}
    \caption{Example 2 from \BenchmarkName~showing reasoning correction.}
    \label{fig:example2}
\end{figure*}

\begin{figure*}[t]
    \centering
    \includegraphics[width=0.95\textwidth,keepaspectratio]{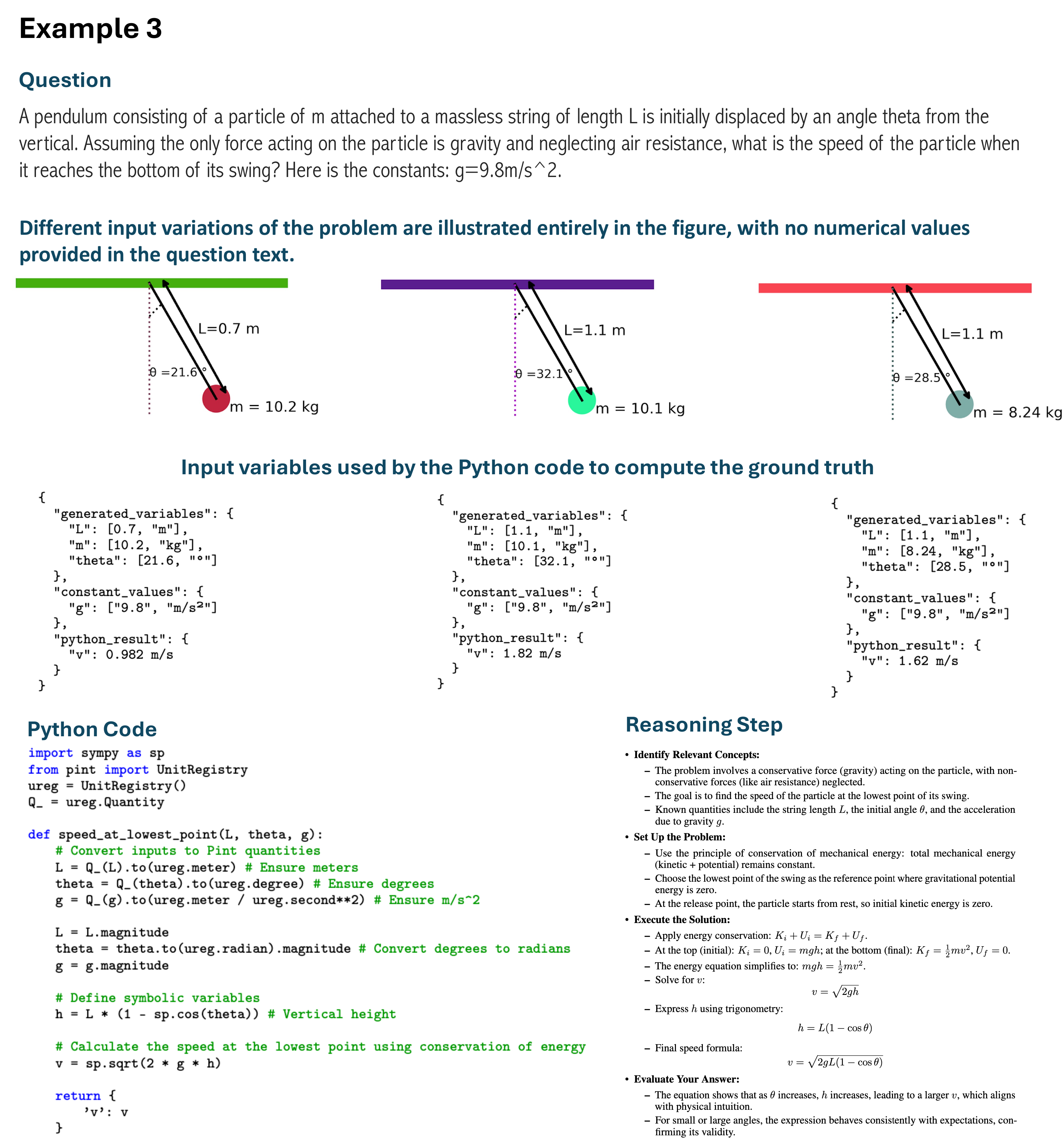}
    \caption{Example 3 from \BenchmarkName~demonstrating multimodal synthesis.}
    \label{fig:example3}
\end{figure*}

% Use figure* for spanning two columns in two-column documents

\subsection{Python Code Scaffolding}
An excerpt of the prompt used to guide the agent in generating fully generalized Python functions and paraphrasing of the problem is shown in \ref{python_scaffolding}.

\begin{figure*}[!htbp]
\begin{tcolorbox}[neuripsbox, title=Prompt Template for Python Code Synthesis, coltitle=black, fonttitle=\bfseries]

\tiny % Set very small font size

Given a problem, perform the following tasks:

\textbf{Generate a Python Function:}
\begin{itemize}
  \item The function should take the provided \texttt{Inputs} as arguments.
  \item It must return only the specified \texttt{Outputs} variables.
  \item Using the given \texttt{Inputs} and the solution, write a Python function that solves the problem and returns the required \texttt{Outputs}.
\end{itemize}

\textbf{Function Requirements:}
\begin{itemize}
  \item Name the function appropriately based on the context of the problem.
  \item The function should take only the identified \texttt{Inputs} variables as arguments.
  \item Use \textbf{SymPy} for symbolic calculations (e.g., algebraic manipulations, solving equations, simplifying expressions).
  \item Use \textbf{Pint} for all unit handling and conversions.
    \begin{itemize}
        \item Apply units to variables using Pint (e.g., \texttt{v0 = Q\_(v0).to(ureg.meter / ureg.second)}) to ensure correct units.
        \item Extract magnitudes using \texttt{.magnitude} before performing numerical calculations.
        \item \textbf{Do not manually convert units} (e.g., avoid converting cm to m manually; always use Pint and then get the magnitude).
    \end{itemize}
  \item Preserve symbolic calculations where possible.
  \item Utilize SymPy functions such as \texttt{solve}, \texttt{dsolve}, \texttt{integrate}, \texttt{diff}, \texttt{simplify}, \texttt{expand}, and \texttt{factor}.
  \item \textbf{IMPORTANT:} Ensure that your Python function is fully generalized—do not hardcode any specific variables. All values should be dynamically derived from the provided input parameters.
  \item If your Python code includes symbolic expressions such as \texttt{sp.sin}, \texttt{sp.pi}, or \texttt{sp.cos}, make sure to apply \texttt{.evalf()} so that when variables are substituted, the output is numerical rather than symbolic.
  \item Always use fully qualified names such as \texttt{sp.cos}, \texttt{sp.sin}, or \texttt{sp.pi} instead of importing symbols directly.
\end{itemize}

\textbf{Python Function Output:}
\begin{itemize}
  \item The function should return a \textbf{dictionary} where:
    \begin{itemize}
      \item \textbf{Keys} correspond to the specified \texttt{Outputs} variable names.
      \item \textbf{Values} are lists of the corresponding parameters.
    \end{itemize}
\end{itemize}

\textbf{Output Format:}
\begin{itemize}
  \item Provide the final output in \textbf{JSON format}.
\end{itemize}

\textbf{Output Format (JSON):}
\begin{verbatim}
import sympy as sp
import numpy as np
from pint import UnitRegistry
ureg = UnitRegistry()
Q_ = ureg.Quantity
def solution(Inputs):
  #### Steps to solve the problem
  return {"ans_1": [ans_1, "unit_1"], "ans_2": [ans2, "unit_2"]}
  # Only return the required Outputs variables
\end{verbatim}

\textbf{Example Output:} for Inputs = [$v_0$, $g$, $y_{\text{target}}$] and Outputs = ['time\_to\_target', 'max\_height']
\begin{verbatim}
import sympy as sp
from pint import UnitRegistry
ureg = UnitRegistry()
Q_ = ureg.Quantity

def projectile_motion(v0, g, y_target):
    # Convert inputs to Pint quantities
    v0 = Q_(v0).to(ureg.meter / ureg.second)  # Ensure m/s
    g = Q_(g).to(ureg.meter / ureg.second**2)  # Ensure m/s²
    y_target = Q_(y_target).to(ureg.meter)  # Ensure meters

    v0 = v0.magnitude
    g = g.magnitude
    y_target = y_target.magnitude

    # Define symbolic variables
    t = sp.Symbol('t', real=True, positive=True)

    # (1) Solve for time to reach y_target
    y = v0 * t - (1/2) * g * t**2  # Motion equation
    time_solutions = sp.solve(sp.Eq(y, y_target), t)

    # Filter out negative time values
    time_to_target = [sol.evalf() for sol in time_solutions if sol.is_real and sol > 0]

    # (2) Calculate maximum height
    t_max = v0 / g  # Time to reach max height (v = 0)
    y_max = v0 * t_max - (1/2) * g * t_max**2

    return {
        'time_to_target': time_to_target,
        'max_height': y_max
      }
\end{verbatim}
\end{tcolorbox}
\end{figure*}\label{python_scaffolding}

\begin{figure*}[!htbp]
\begin{tcolorbox}[neuripsbox, title=Prompt Template for Paraphrasing I, coltitle=black, fonttitle=\bfseries]

\tiny % Very small font size

Paraphrase the given generic question and solution while ensuring clarity, creativity, and structured refinement. Follow these guidelines:

\textbf{Important:}
The list of symbols ending with \texttt{\_numerical\_value} (called \textit{immutable\_symbols}) must remain unchanged in both question and solution. Do not add new symbols ending with \texttt{\_numerical\_value}.

\textbf{Paraphrasing Guidelines:}
\begin{itemize}
  \item \textbf{Preserve Parameter Names:} Keep all variable names ending with \texttt{\_numerical\_value} unchanged. Do not add units.
  \item \textbf{Enhance Readability \& Conciseness:} Restructure sentences for clarity and flow; remove redundancies while keeping all key details.
  \item \textbf{Refine References \& Equations:} Remove mentions of equation numbers or figure references; present equations naturally.
  \item \textbf{Introduce Meaningful Variations:} Modify the scenario creatively (e.g., change "car" to "truck") while preserving core concepts, methods, and solution approach.
  \item \textbf{Maintain Structural Integrity:} Rewrite the solution using the provided problem-solving template, mirroring original reasoning but in reworded form.
  \item \textbf{Make Questions More Abstract:} Ensure the paraphrased question feels like a new problem with improved clarity.
\end{itemize}

\textbf{Output Format:}
Return a JSON object with keys:
\begin{verbatim}
{
"Paraphrase_Question": "Paraphrased question with immutable symbols intact.",
"Paraphrase_Calculation": "Paraphrased solution following the problem-solving template."
}
\end{verbatim}

\textbf{Example Template:}

\textit{Problem Solving Steps:}
\begin{itemize}
  \item Identify Relevant Concepts
  \item Set Up the Problem
  \item Execute the Solution
  \item Evaluate Your Answer
\end{itemize}

\textbf{Note:} For all examples, symbols ending with \texttt{\_numerical\_value} are never changed.

\end{tcolorbox}
\end{figure*}

\begin{figure*}[!htbp]
\begin{tcolorbox}[neuripsbox, title=Prompt Template for Paraphrasing II, coltitle=black, fonttitle=\bfseries]

\tiny % Very small font size

Paraphrase the given generic question and solution while ensuring clarity, creativity, and structured refinement. Follow these updated guidelines to enhance the quality of the paraphrased output:

\textbf{Most Important:}
The list of symbols ending with \texttt{\_numerical\_value} (referred to as \textit{immutable\_symbols}) must remain unchanged in both the question and solution. Do \textbf{not} modify, rename, or introduce any new variables ending with \texttt{\_numerical\_value}.

\textbf{Paraphrasing Guidelines:}
\begin{itemize}
  \item \textbf{Preserve Parameter Names:} All variables ending with \texttt{\_numerical\_value} must remain unaltered. Do not append, rename, or introduce new ones. \textbf{Do not include units.}

  \item \textbf{Enhance Readability \& Conciseness:} Improve sentence structure for clarity and smooth flow. Eliminate redundancies while preserving all essential information. The text should be natural, precise, and easy to follow.

  \item \textbf{Refine References \& Equations:} Remove all references to figure numbers or equations unless explicitly present in the original. Present equations naturally within the flow of the text without reference annotations.

  \item \textbf{Introduce Meaningful Variations:} Modify the context while preserving the underlying mathematical or physical principles. For example, change "car" to "truck" or "rabbit" to "fox." Be creative but maintain logical and conceptual equivalence. Ensure that the original equations and methodology remain applicable.

  \item \textbf{Maintain Structured Reasoning:} Paraphrased solutions must follow the problem-solving template outlined below. Keep the logical structure of the original reasoning intact, but reword and reframe it for improved clarity.

  \item \textbf{Make the Questions More Abstract:} Rephrase the question to sound more generalized or conceptually framed. Avoid overly specific or concrete phrasing if not required.
\end{itemize}

\textbf{Problem Solving Template:}
\begin{itemize}
  \item \textbf{Read and Understand:} Read the problem thoroughly and ensure full comprehension.
  \item \textbf{Identify Objects and Quantities:} Determine the key objects, knowns, and unknowns. Translate language into mathematical or physical symbols as needed.
  \item \textbf{Apply Relevant Principles:} Identify the laws or equations that govern the situation and plan a logical approach.
  \item \textbf{Select and Apply Equations:} Choose appropriate equations, solve symbolically, and then substitute numerical values.
  \item \textbf{Calculate and Verify:} Perform the computation and verify the accuracy of the result.
  \item \textbf{Check Units:} Ensure dimensional consistency (if applicable), but do not include explicit units in the paraphrased text.
\end{itemize}

\textbf{Output Format:}
Return your output in the following JSON format:

\begin{verbatim}
{
  "Paraphrase_Question": "Paraphrased question with immutable symbols intact.",
  "Paraphrase_Calculation": "Paraphrased solution following the structured problem-solving template."
}
\end{verbatim}

\textbf{Example:}

\begin{verbatim}
{
  "Paraphrase_Question": "Consider a mathematical function f(x) = c_numerical_value x + d_numerical_value defined over a specific range from a_numerical_value to b_numerical_value. What is the mean value of this function within the given interval, knowing that the area under its curve can be geometrically represented?",
  "Paraphrase_Calculation": "To solve this, start by identifying the linear function and the interval it spans. This function forms a trapezoidal area under the curve between x = a and x = b. Begin by evaluating the function at the endpoints: f(a) = c \cdot a + d and f(b) = c \cdot b + d. Use the trapezoid area formula A = (1/2) \cdot (\text{base}) \cdot (\text{sum of parallel sides}), where the base is (b - a) and the parallel sides are f(a) and f(b). Plugging in the values gives: \int_{a}^{b} (cx + d) \, dx = (1/2) \cdot (b - a) \cdot (f(a) + f(b)) = (1/2) \cdot (b - a) \cdot (c(a + b) + 2d). To find the average value, divide the integral by the length of the interval: (1 / (b - a)) \cdot \int_{a}^{b} (cx + d) \, dx = (1/2) \cdot (c(a + b) + 2d). Hence, the average value of the function over [a, b] is (1/2) \cdot (c(a + b) + 2d)."
}
\end{verbatim}

\end{tcolorbox}
\end{figure*}

\begin{figure*}[!htbp]
\begin{tcolorbox}[neuripsbox, title=Prompt Template for Paraphrasing III, coltitle=black, fonttitle=\bfseries]

\tiny % Very small font size

Paraphrase the given generic question and solution while ensuring clarity, creativity, and structured refinement. Follow these enhanced guidelines to improve the quality and instructional clarity of the paraphrased output.

\textbf{Most Important:}
The list of symbols ending with \texttt{\_numerical\_value} (called \textit{immutable\_symbols}) must remain unchanged in both the question and solution. Do not modify existing ones or introduce new symbols with this suffix.

\textbf{Paraphrasing Guidelines:}
\begin{itemize}
  \item \textbf{Preserve Parameter Names:}
  Keep all variable names ending with \texttt{\_numerical\_value} intact. \textbf{Do not change these symbols or include units} in either the question or the solution.

  \item \textbf{Enhance Readability \& Conciseness:}
  Restructure sentences to improve clarity and flow. Eliminate redundant phrases while preserving all necessary information. Ensure the paraphrased text is engaging, readable, and logically organized.

  \item \textbf{Refine References \& Equations:}
  Do not refer to figure numbers or equation numbers unless they are explicitly included in the original text. Write equations as part of the natural narrative of the solution.

  \item \textbf{Introduce Meaningful Variations:}
  Change the context of the problem while maintaining the core concept, underlying method, and the validity of the original equations. For example, replace "asteroid" with "satellite", or "Earth" with "a planet". Ensure logical and scientific consistency in the new context.

  \item \textbf{Maintain Structured Problem Solving:}
  Rewrite the solution using the specific problem-solving strategy outlined below. The structure should mirror the logical approach of the original solution, but with new wording and context.

  \item \textbf{Make the Questions More Abstract:}
  Avoid overly specific phrasing unless necessary. Frame the problem in a slightly more general or conceptual manner to give it a fresh presentation.
\end{itemize}

\textbf{Structured Problem Solving Template:}
\begin{itemize}
  \item \textbf{List Known and Unknowns:}
  Record all given parameters and identify what is to be solved.

  \item \textbf{Symbolic Solutions:}
  Solve the problem using algebra before plugging in numerical values.

  \item \textbf{Units and Dimensions:}
  Ensure that the dimensional units used are consistent and correct throughout the process. Do not write units explicitly in the output.

  \item \textbf{Check Special Cases:}
  Evaluate limiting or boundary cases (e.g., zero initial speed) to validate that the solution behaves reasonably. Conduct a basic sanity check on the final expression or value.
\end{itemize}

\textbf{Output Format:}
Return your final output in the following JSON format:

\begin{verbatim}
{
  "Paraphrase_Question": "Paraphrased question that introduces contextual variation but keeps all
                         immutable symbols unchanged.",
  "Paraphrase_Calculation": "Step-by-step solution restructured using the provided problem-solving template."
}
\end{verbatim}

\textbf{Example:}

\begin{verbatim}
{
  "Paraphrase_Question": "A small satellite is launched directly toward a planet and has an initial speed of
    v_i_numerical_value when it is located 10 R_E_numerical_value from the planet’s center. Ignoring any
    atmospheric drag or resistance, determine the satellite's speed v_f upon reaching the planet’s surface.
    The following constants are provided: {\"G\": G_numerical_value, \"M\": M_numerical_value,
    \"R_E\": R_E_numerical_value}",

  "Paraphrase_Calculation": "To determine the satellite’s speed when it reaches the planet's surface, we follow
    these structured problem-solving steps:\n\n

    - List Known and Unknowns:\n
      - Known: Initial speed = v_i_numerical_value, initial distance = 10 R_E_numerical_value,
        final distance = R_E_numerical_value, constants G = G_numerical_value and M = M_numerical_value.\n
      - Unknown: Final speed v_f.\n\n

    - Symbolic Solutions:\n
      - Since there are no dissipative forces, total mechanical energy is conserved: \n
        K_initial + U_initial = K_final + U_final\n
      - Let m be the satellite's mass. Kinetic energy is (1/2)mv^2, and gravitational potential energy is
        -GMm/r.\n
      - Substituting values:\n
        (1/2)mv_f^2 - GMm/R_E = (1/2)mv_i^2 - GMm/(10R_E)\n
      - Solving algebraically for v_f:\n
        v_f^2 = v_i^2 + (2GM/R_E)(1 - 1/10)\n\n

    - Units and Dimensions:\n
      - Check that all constants and distances are compatible in terms of dimensions.\n\n

    - Check Special Cases:\n
      - If v_i_numerical_value = 0, the equation still works, giving a result based purely on gravitational
        attraction.\n
      - Confirm that the final velocity value is physically reasonable given the context.\n\n

    Thus, the satellite's final speed v_f can be computed using the above expression."
}
\end{verbatim}

\end{tcolorbox}
\end{figure*}

\end{document}